\documentclass[final,5p,times,twocolumn]{elsarticle}
\let\today\relax
\makeatletter
\def\ps@pprintTitle{%
    \let\@oddhead\@empty
    \let\@evenhead\@empty
    \def\@oddfoot{\footnotesize\itshape
         {} \hfill\today}%
    \let\@evenfoot\@oddfoot
    }
\makeatother

\usepackage{amssymb}
\usepackage{graphicx} 

\usepackage{array}

\usepackage{algorithm}
\usepackage{algpseudocode}
\usepackage{latexsym}
\usepackage{multirow}
\usepackage{amsmath}
\usepackage{comment}
\usepackage{tabularx}
\usepackage{natbib}
\usepackage{float}
\usepackage{subcaption}
\usepackage{pdflscape}
\usepackage{graphicx}
\usepackage{longtable}
\usepackage{adjustbox}
\usepackage[table]{xcolor}
\usepackage{lscape}
\usepackage{pdfpages}
\usepackage{afterpage}
\usepackage{capt-of}

\usepackage{soul}

\usepackage{color}
\soulregister{\cite}{7}
\soulregister{\citep}{7}
\soulregister{\citet}{7}
\soulregister\Hl{7}
\soulregister{\ref}{1}
\soulregister{\itemsize}{1}
\soulregister{\textbf}{1}
\soulregister{\footnote}{1}

\usepackage{booktabs}  

\begin{document}

\begin{frontmatter}

\title{Integrating Probabilistic Trees and Causal Networks for Clinical
and Epidemiological Data}

\author{Sheresh Zahoor\corref{cor1}\fnref{label1}}
 \ead{sheresh.zahoor@mycit.ie}
 
 \cortext[cor1]{Corresponding author}
\affiliation[label1]{organization={Munster Technological University},
             addressline={Rossa Ave, Bishopstown},
            city={Cork},
            country={Ireland}}

 

\author[label2]{Pietro Liò}

\affiliation[label2]{organization={Department of Computer Science and Technology, University of Cambridge},
           addressline={The Old Schools, Trinity Ln}, 
            city={Cambridge},
            country={United Kingdom}}

\author[label4]{Gaël Dias}
\affiliation[label4]{organization={Université Caen Normandie, ENSICAEN, CNRS, Normandie Univ, GREYC UMR6072, F-14000 Caen, France},
           addressline={Boulevard du Maréchal Juin}, 
            city={Caen},
            country={France}}

\author[label3]{Mohammed Hasanuzzaman}
\affiliation[label3]{organization={The School of Electronics, Electrical Engineering and Computer Science, Queens University Belfast},
           addressline={University Road}, 
            city={Belfast},
            country={United Kingdom}}

\begin{abstract}
Healthcare decision-making requires not only accurate predictions but also insights into how factors influence patient outcomes. While traditional machine learning (ML) models excel at predicting outcomes, such as identifying high-risk patients, they are limited in addressing "what if" questions about interventions. This study introduces the Probabilistic Causal Fusion (PCF) framework, which integrates Causal Bayesian Networks (CBNs) and Probability Trees (PTrees) to extend beyond predictions. PCF leverages causal relationships from CBNs to structure PTrees, enabling both the quantification of factor impacts and the simulation of hypothetical interventions. PCF was validated on three real-world healthcare datasets i.e. MIMIC-IV, Framingham Heart Study, and Diabetes—chosen for their clinically diverse variables. It demonstrated predictive performance comparable to traditional ML models while providing additional causal reasoning capabilities. To enhance interpretability, PCF incorporates sensitivity analysis and SHapley Additive exPlanations (SHAP). Sensitivity analysis quantifies the influence of causal parameters on outcomes such as Length of Stay (LOS), Coronary Heart Disease (CHD), and Diabetes, while SHAP highlights the importance of individual features in predictive modeling. By combining causal reasoning with predictive modeling, PCF bridges the gap between clinical intuition and data-driven insights. Its ability to uncover relationships between modifiable factors and simulate hypothetical scenarios provides clinicians with a clearer understanding of causal pathways. This approach supports more informed, evidence-based decision-making, offering a robust framework for addressing complex questions in diverse healthcare settings.

\end{abstract}


\end{frontmatter}


\section{Introduction}\label{sec1}

Effective healthcare requires not just accurate predictions but also a deeper understanding of the factors that drive patient outcomes. While traditional machine learning (ML) models are proficient at identifying patterns and forecasting risks—such as predicting which patients are more likely to develop a condition—they often fall short in evaluating how specific interventions might alter these outcomes. This predictive focus limits their utility in addressing causal questions, such as estimating the impact of treatments or other modifiable factors on patient trajectories. To bridge this gap, there is a growing need for approaches that go beyond prediction, enabling the quantification of causal effects and simulation of intervention outcomes.

Causal ML addresses these gaps by estimating treatment effects and answering counterfactual questions, such as "How would a patient’s outcome change if a different treatment were administered?" Unlike traditional ML, which focuses on correlations, causal ML is built on the foundation of causal inference \cite{reason:Pearl09a}, enabling a deeper understanding of relationships and supporting evidence-based decision-making \cite{sanchez2022causal}.For instance, traditional ML might predict a patient’s likelihood of developing diabetes \cite{kopitar2020early}, but causal ML can estimate how that likelihood would change under specific interventions, such as a lifestyle modification or a new medication \cite{feuerriegel2024causal}. These capabilities are particularly valuable in healthcare, where understanding cause-effect relationships is critical for developing targeted interventions \cite{prosperi2020causal}.

To address these challenges, we propose the Probabilistic Causal Fusion (PCF) framework, which integrates Causal Bayesian Networks (CBNs) and ensembles of Probability Trees (PTrees) to bridge the gap between prediction and causality. CBNs use a directed acyclic graph (DAG) structure to model dependencies and causal pathways in healthcare data \cite{sanchez2022causal, prosperi2020causal, rajput2022causal, ambags2023assisting}, while PTrees offer an intuitive representation of probabilistic relationships, enabling reasoning about uncertainty and interactions between variables \cite{genewein2020algorithms}. By combining these complementary approaches, PCF provides a powerful framework for understanding the relationships between factors and their impact on patient outcomes.

While Randomised Controlled Trials (RCTs) remain the gold standard for establishing causal relationships, they are often resource-intensive and ethically challenging. PCF offers an alternative by leveraging observational healthcare data to uncover causal relationships, quantify treatment effects, and simulate hypothetical interventions. Unlike traditional ML models, which focus on predicting outcomes based on correlations, PCF enables clinicians to explore the causal pathways underlying these outcomes and assess the effects of specific interventions.

The hierarchical structure of PCF enhances its clinical utility by leveraging the strengths of both CBNs and PTrees. CBNs investigate dependencies among morbidities and assess temporal relationships, while PTrees quantify the strength of these relationships across patient cohorts. This approach allows for the seamless integration of diverse data sources, such as clinical bedside information, into a unified framework for decision support. By using the causal knowledge derived from CBNs to structure the PTrees, PCF ensures that probabilistic dependencies reflect the underlying causal relationships in the data. This not only reduces the reliance on subjective domain expertise but also ensures consistency between the data and the model.

Traditionally, constructing PTrees has been an iterative process reliant on domain experts to define variable orderings \cite{ambags2023assisting}, which introduces challenges such as: \begin{enumerate} 
\item Subjectivity: Expert knowledge may be incomplete or biased, leading to suboptimal structures. 
\item Inefficiency: Manual construction is time-intensive, especially for complex datasets. 
\item Data Inconsistency: Sole reliance on expert-defined structures can overlook key relationships present in the data. 
\end{enumerate} 
By employing an ensemble learning approach, PCF mitigates these limitations, offering a robust and generalisable solution that reduces overfitting and enhances performance.

\begin{figure*}
\centering
    \includegraphics[width=\textwidth]{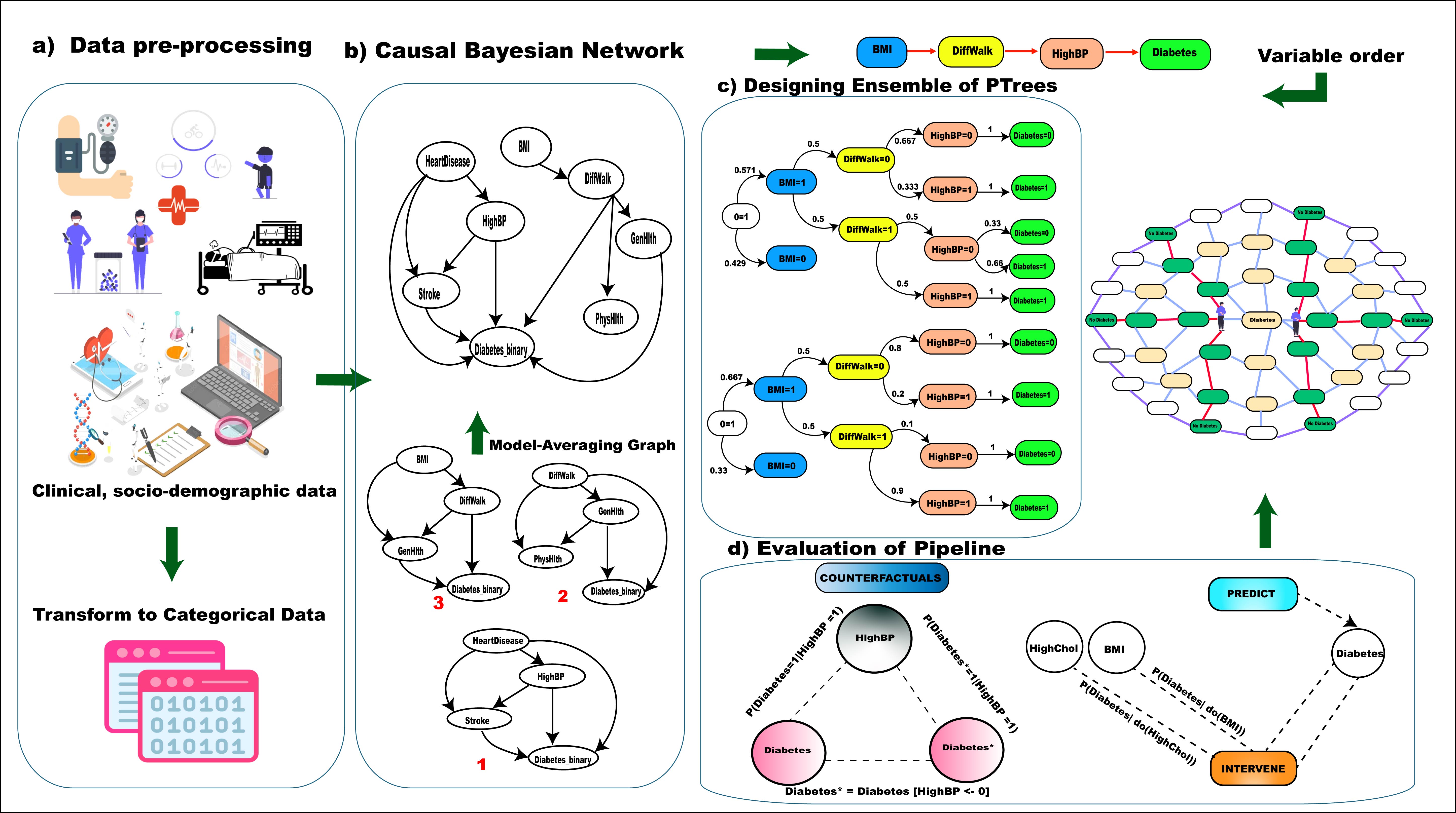}
    \caption{Different steps involved in the PCF framework. The first module addresses data pre-processing to shape the input required for the CBN. The next module involves generating individual CBNs and creating a model-averaging graph. Subsequently, the ensemble of PTrees is developed based on the variable order from the model-averaging graph. The final module involves evaluating the overall performance of PCF.}
    \label{fig:illustration}
\end{figure*}

Beyond its modeling capabilities, PCF also establishes a centralised causal knowledge repository, fostering collaboration among healthcare institutions. Pre-trained PCF models, along with their underlying causal structures, can be shared across institutions, enabling smaller organisations to leverage collective expertise and refine models using their local data. This collaborative ecosystem ensures continuous improvement and supports the development of generalisable causal models that can address diverse healthcare challenges. Figure \ref{fig:illustration} summarises the steps involved in the proposed PCF framework.

In this work, we leverage the PCF framework to support prediction, intervention, and counterfactual analysis in three distinct clinical contexts. First, we aim to predict and identify factors associated with the length of stay in the Intensive Care Unit (ICU). Second, the framework is applied to assess the risk of Chronic Heart Disease (CHD) and investigate potential modifiable factors that could mitigate its progression. Finally, we utilise the framework to predict the risk of diabetes and analyse the influence of various risk factors on its onset.

The main contributions of this work are summarised as follows: \begin{itemize} 
\item We propose a framework, Probabilistic Causal Fusion (PCF), that combines Causal Bayesian Networks (CBNs) with ensembles of Probability Trees (PTrees) to enable predictions, interventions, and counterfactual analysis in healthcare. This integration addresses the limitations of traditional PTree construction, such as reliance on domain expertise for variable ordering, by leveraging causal relationships identified through CBNs. 
\item The use of an ensemble of PTrees improves predictive performance and robustness. By balancing the trade-off between bias and variance, the ensemble approach mitigates risks of overfitting or underfitting and enhances generalisability. 
\item Methodological refinements are introduced in computing transition probabilities within the PTree framework. Specifically, data is partitioned using empirical marginal probabilities, while causal relationships derived from CBNs inform split decisions, enabling more data-driven and effective model construction. 
\item The applicability and utility of the PCF framework are demonstrated through its application to multiple real-world healthcare datasets. 
\end{itemize}

The structure of the paper is as follows. Section \ref{Relevant} provides an overview of the relevant literature, establishing the context and motivation for this work. Section \ref{method} describes the proposed framework in detail, including the steps involved in its construction and implementation. Section \ref{casestudies} presents the application of the framework to three distinct clinical case studies, while Section \ref{results} discusses the results obtained from these applications. Finally, Section \ref{conclusion} provides conclusive remarks and directions for future work.

\section{Relevant Works} \label{Relevant}
\subsection{Traditional ML vs. Causal ML}
Traditional ML has become a cornerstone in healthcare for tasks such as risk prediction, patient stratification, and outcome forecasting \cite{chen2023custom, usman2023systematic}. These models are highly effective at identifying patterns and correlations, enabling predictions such as the likelihood of disease onset or hospital readmissions \cite{mennickent2022machine,ahsan2022machine, ma2024hr}. However, traditional ML lacks the ability to answer counterfactual questions or estimate treatment effects, as it is inherently focused on predictive accuracy rather than causal reasoning \cite{yao2021survey,blumberg2016causal}.

In contrast, causal ML seeks to address "what if" questions by estimating the causal effect of interventions on outcomes. For instance, rather than simply predicting the probability of diabetes onset, causal ML can assess how this probability might change if a patient adopts a new medication or lifestyle modification \cite{belthangady2022causal, richens2020improving}. These capabilities enable healthcare providers to explore counterfactual scenarios and make data-driven decisions that go beyond prediction to inform treatment planning and resource allocation.

Causal ML requires additional considerations compared to traditional ML, such as the need to account for unobservable outcomes and confounding variables. For example, the "fundamental problem of causal inference" \cite{holland1986statistics, reason:Pearl09a} states that only factual outcomes under a given treatment are observed, while counterfactual outcomes remain unobserved. This poses challenges in estimating treatment effects and necessitates assumptions such as the absence of unmeasured confounders to ensure unbiased estimates. Additionally, causal ML models must account for the interconnectedness of variables, as interventions on one factor may influence others. For example, a smoking cessation intervention may indirectly affect diabetes risk through changes in body mass index (BMI), requiring a causal framework to model such dependencies.

\subsection{Applications of CBNs in Healthcare}
CBNs are a widely used tool in causal ML for modeling relationships among variables through a directed acyclic graph (DAG) structure. CBNs allow for the incorporation of prior knowledge and probabilistic reasoning, making them particularly effective for understanding complex dependencies in healthcare data. For instance, Rajendran et al. \cite{rajendran2020predicting} employed CBNs to integrate risk factor analysis in breast cancer research, facilitating early detection and risk stratification. Shahmirzalou et al. \cite{shahmirzalou2023survival} applied CBNs to recurrent breast cancer data to predict survival outcomes and guide personalised treatment strategies. Similarly, Jang et al. \cite{jang2024estimating} leveraged CBNs to model risks associated with radiation therapy, offering support for personalised oncology care.

CBNs have also been applied to explore relationships between cardiovascular risk factors and related conditions, as demonstrated by Ordovas et al. \cite{ordovas2023bayesian}. These examples highlight the versatility of CBNs in identifying risk factors, modeling disease progression, and simulating the effects of interventions. However, while CBNs excel at representing causal relationships, their construction often requires significant domain expertise and computational resources, which may limit scalability.

\subsection{Probability Trees (PTrees) in Causal Modeling}
Probability Trees (PTrees) offer an intuitive and sequential representation of probabilistic relationships. Grounded in probability theory, PTrees use a tree structure where each node represents a probabilistic event and branches correspond to the conditional probabilities of subsequent events \cite{genewein2020algorithms}. This structure is particularly useful for modeling complex decision-making processes in healthcare, where uncertainty and sequential dependencies play a significant role.

Despite their simplicity and expressiveness, PTrees have received relatively less attention in the machine learning literature compared to CBNs or structural causal models. Ambags et al. \cite{ambags2023assisting} proposed a hybrid approach combining probabilistic fuzzy decision trees with causal reasoning, applying it in two medical case studies to demonstrate its potential for real-world applications. However, traditional PTree construction often relies on domain expertise for defining variable order, which can introduce subjectivity, inefficiency, and inconsistencies between expert-defined structures and data-driven insights.

By integrating PTree ensembles with causal insights derived from CBNs, frameworks like Probabilistic Causal Fusion (PCF) aim to address these limitations. The use of ensembles mitigates overfitting risks, while the causal order provided by CBNs ensures that the modeled relationships align with the underlying data-generating process.

\subsection{Advances in Causal ML for Healthcare}
Recent advances in causal ML demonstrate its potential for improving healthcare decision-making by estimating treatment effects and simulating counterfactual scenarios. These methods have been applied to a variety of clinical challenges, from estimating the effects of medication adherence on diabetes progression to predicting survival probabilities under different oncology treatment plans \cite{prosperi2020causal, sanchez2022causal}. By integrating causal ML techniques into clinical workflows, researchers aim to bridge the gap between prediction and actionable insights, enabling data-driven strategies for improving patient outcomes.

However, causal ML comes with its own challenges. Assumptions about unmeasured confounding, model scalability, and the reliability of observational data remain critical concerns \cite{dahabreh2024causal}. Addressing these limitations through frameworks like PCF, which combine causal reasoning with robust predictive modeling, represents an important step toward leveraging causal ML in diverse healthcare contexts.
\section{Methodology-PCF framework}\label{method}

This study introduces PCF, a novel framework combining the strengths of CBNs and PTrees for enhanced predictions, interventions, and counterfactual analyses. Traditionally, PTree construction involves selecting and ordering features based on domain expertise (deduction) and setting transition probabilities from empirical data (induction). This process is subjective and can impact model performance. Our framework leverages CBNs to autonomously determine the variable order, streamlining PTree construction and enhancing efficiency and accuracy. Additionally, forming an ensemble of multiple PTrees improves predictive performance by mitigating overfitting and underfitting through balancing bias and variance across diverse data subsets.
\subsection{CBN Construction}
CBNs are a powerful framework for elucidating the causal relationships among variables in complex datasets. In this phase, we utilised CBNs to reveal the causal architecture of selected variables and ascertain the optimal sequence in which they exert influence on the outcome variable \( Y \). This sequencing is crucial for the subsequent development of a proficient Probability Tree (PTree).
Given the  set of variables \( V = \{X_1, X_2, \ldots, X_n\} \) and the outcome variable \( Y \), we construct a Causal Bayesian Network (CBN) \( G \) such that:

\begin{itemize}
    \item \( G \) represents the causal dependencies among the variables in \( V \) and \( Y \).
    \item We identify a topological ordering of the variables that reflects the causal influence sequence leading to \( Y \).
\end{itemize}
We employed the "target variable" knowledge approach from the Bayesys \cite{anthony} open-source Bayesian Network Structure Learning (BNSL) system to prioritise structure learning to elucidate a greater number of parent nodes, revealing potential causal factors for the variable of interest i.e. \( Y \)
\subsubsection{Methodological Steps}
\begin{itemize}
    \item[1.] \textbf{Structure Learning}:  For each algorithm \( A_i \), we aim to find the optimal network structure \( G_i \) that maximises a given scoring function \( S \). The algorithms used include Hill Climbing (HC) \cite{heckerman1995learning,hc}, TABU \cite{bouckaert1995bayesian}, SaiyanH \cite{saiyanh}, Model-Averaging Hill-Climbing (MAHC) \cite{constantinou2022effective}, and Greedy Equivalence Search (GES) \cite{ges} The optimal structure \( G_i \) is determined by:

\[
G_i = \arg\max_G S(G \mid \text{data})
\]
These algorithms were chosen for their capability to incorporate the target variable during model construction and to handle diverse knowledge approaches, including direct relationships and forbidden edges \cite{constantinou2023impact}.
\item[2.] \textbf{Model Averaging}: To address the biases and limitations of individual algorithms, we adopted the model-averaging strategy implemented in Bayesys \cite{anthony} to combine multiple structures \( G_i \) into an averaged graph \( G_{\text{avg}} \). The following steps outline this process:
\begin{itemize}
\item[a.] \textbf{Directed Edges}: Add directed edges \( e = (u, v) \) to \( G_{\text{avg}} \) starting with the edges that occur most frequently across input graphs, ensuring no cycles are formed:

\[
e \in G_{\text{avg}} \text{ if freq}(e) > \text{threshold and no cycle}
\]

If adding an edge \( e \) would create a cycle, reverse the edge:

\[
e \rightarrow e^{-1} \text{ if } e \text{ forms a cycle}
\]

\item[b.] \textbf{Undirected Edges}: Add undirected edges \( e = \{u, v\} \) to \( G_{\text{avg}} \), skipping those already added as directed edges:

\[
e \in G_{\text{avg}} \text{ if freq}(e) > \text{threshold}
\]

\item[c.] \textbf{Cycle Handling}:Add directed edges from the cycle-inducing edge set \( C \):

\[
e \in G_{\text{avg}} \text{ if } e \in C \text{ and occurs frequently}
\]
By employing this model-averaging approach, we create a single Directed Acyclic Graph (DAG) that consolidates insights from all considered structure learning algorithms. This enhances the robustness and reliability of the learnt graph, mitigating algorithmic biases and addressing limitations in edge orientation from observational data. 
\end{itemize}

\item[3.] \textbf{Topological Sorting}:
Determine a topological order \( \pi \) of the nodes in \( G_{\text{avg}} \). This ordering is crucial for ensuring that the nodes are processed in a sequence that respects the causal dependencies:

\[
\pi = \text{topological\_sort}(G_{\text{avg}})
\]

\end{itemize}
The sorted order \( \pi \) is then used in the construction of PTrees, facilitating their development and enhancing their predictive capabilities.

\subsubsection{Sensitivity Analysis} 
Sensitivity analysis is performed to assess the responsiveness of nodes to changes in their parent and ancestor nodes \cite{kjaerulff2013making}. For a node \( X_i \) with parent \( X_j \), sensitivity \( S \) is defined as:

\[
S = \frac{\partial P(X_i)}{\partial \theta_{X_j}}
\]

where \( \theta_{X_j} \) represents the parameters in the Conditional Probability Table (CPT) of \( X_j \). High sensitivity indicates that small changes in \( \theta_{X_j} \) result in significant changes in the posterior distribution of \( X_i \), suggesting a strong dependency. Conversely, low sensitivity implies that large changes in \( \theta_{X_j} \) have minimal impact on \( X_i \)'s distribution, indicating a weak dependency. 

The posterior probability \( T \) of the selected state of the target node, given the parameter \( p \), is represented by the following general linear rational functional form:

\[
T = \frac{a \cdot p + b}{c \cdot p + d}
\]

The sensitivity analysis algorithm calculates the coefficients \( a, b, c, \) and \( d \). The derivative, which is the basic measure of sensitivity, is given by:

\[
D = \frac{a \cdot d - b \cdot c}{(c \cdot p + d)^2}
\]

The denominator is positive, indicating that the sign of the derivative is constant for all values of \( p \). By substituting 0 and 1 for \( p \) (noting that \( p \) is a probability), we can calculate the range within which the posterior will change if \( p \) is modified across its entire range, defined by:

\[
p_1 = \frac{b}{d}
\]

\[p_2 = \frac{a + b}{c + d}\]

Sensitivity analysis is crucial in understanding the stability and robustness of the model. It helps identify the most influential parameters in the network, guiding targeted interventions and enhancing the interpretability of the model. We use the GeNIe BN software \cite{genie} to perform this analysis.

\subsection{Probability Tree Construction}

Based on the CBN framework, we construct Probability Trees (PTrees) as follows:

\subsubsection{Create Tree from Data}
The process commences with Algorithm \ref{alg:create_tree} (\textsc{create\_tree\_from\_data}), which outlines the construction of a PTree from a given dataset and the variable order derived from the previously learned CBN.
\begin{itemize}
    \item[a.] Construct a PTree \( T \) from dataset \( D \) using the variable order \( \pi \) derived from the CBN \( G \).
    \item[b.] \textbf{Root Level:} Partition data based on the prevalent value of the target attribute \( Y \):

    \[
    p(v) = \frac{\text{Count}(v)}{\text{Total Samples}}
    \]

    where:
    \begin{itemize}
        \item \( p(v) \) denotes the initial probability for a given value \( v \) of \( Y \).
        \item \(\text{Count}(v)\) is the number of occurrences of \( v \) in \( Y \).
        \item \(\text{Total Samples}\) is the total number of samples.
    \end{itemize}
This initial partitioning uses empirical marginal probabilities to establish a foundation for the tree structure.

    \item[c.] \textbf{Transition Probabilities:} Calculate transition probabilities using the causal structure:

    \[
    P(X_{\text{current}} \mid X_{\text{parent}}) = \frac{\text{Count}(X_{\text{current}}, X_{\text{parent}})}{\text{Count}(X_{\text{parent}})}
    \]
    where:
\begin{itemize}
    \item \(P(X_{\text{current}} | X_{\text{parent}})\) represents the transition probability of the current attribute given the parent attribute.
    \item \(\text{Count}(X_{\text{current}}, X_{\text{parent}})\) denotes the number of occurrences of the current value given the parent value.
    \item \(\text{Count}(X_{\text{parent}})\) represents the total number of occurrences of the parent value.
\end{itemize}
This context-aware approach leverages the parent-child relationships identified by the CBN, enabling the tree to make more precise and informed splits, thereby potentially enhancing its predictive capabilities.
\item[d.] \textbf{Pruning:} Employ pruning to prevent overfitting by evaluating conditional probabilities against a threshold. Only branches with probabilities exceeding this threshold are retained, ensuring the tree focuses on the most significant splits.
\end{itemize}

\subsubsection{Ensemble Learning}
A key innovation of our proposed method is the use of ensemble learning to build a robust prediction model. This approach enhances the reliability and accuracy of the predictions.
The ensemble learning process within our framework is implemented through the \textsc{ensemble\_probability\_trees} function, which follows a three-step strategy:

\begin{itemize}
    \item[a.] \textbf{Data Splitting:} Divide \( D \) into \( k \) distinct subsets \(\{D_1, D_2, \ldots, D_k\}\).
    \item[b.] \textbf{Diverse PTrees:} Construct a PTree \( T_i \) for each subset \( D_i \) using Algorithm \ref{alg:create_tree} \textsc{create\_tree\_from\_data}:

    \[
    T_i = \textsc{create\_tree\_from\_data}(D_i, \pi)
    \]

    \item[c.] \textbf{Root Node Storage:} Store root nodes of individual PTrees in a list \texttt{ptrees}.
\end{itemize}

\subsubsection{Prediction Process}
To leverage the ensemble of PTrees for making predictions, we follow these steps:
\begin{itemize}
    \item[a.] \textbf{Individual Predictions:} For each PTree \( T_i \), generate a prediction \(\hat{y}_i\) for a given data point using the \text{Predict} function. This step utilises the structure and relationships captured within each tree to produce a prediction:
    
\[
    \hat{y}_i = \text{Predict}(T_i, \text{data point})
    \]
Each tree in the ensemble provides its own prediction, reflecting the diverse insights each tree has gained from its subset of the data. As an integral component of the predictive framework, Algorithm \ref{alg:conditional_probability} calculates the conditional probability of a specified class based on a defined set of input conditions. 
    \item[b.] \textbf{Averaging Predictions:} Combine the predictions from all \( k \) PTrees by calculating the average prediction. This step helps to smooth out individual tree biases and improve overall predictive accuracy:

    \[
    \hat{y}_{\text{avg}} = \frac{1}{k} \sum_{i=1}^{k} \hat{y}_i
    \]
The averaged prediction \(\hat{y}_{\text{avg}}\) provides a consensus estimate, leveraging the collective knowledge of the entire ensemble.

    \item[c.] \textbf{Threshold Classification:} Classify data points based on the averaged prediction using a predefined threshold \( \tau \). This classification step assigns a final class label, determining whether the data point belongs to the positive or negative class:

    \[
    \text{Class} = 
    \begin{cases} 
        \text{Positive} & \text{if } \hat{y}_{\text{avg}} > \tau \\
        \text{Negative} & \text{if } \hat{y}_{\text{avg}} \leq \tau 
    \end{cases}
    \]
The threshold \( \tau \) can be adjusted depending on the specific requirements of the application, allowing for flexible decision-making.
\end{itemize}
By incorporating predictions from this diverse ensemble of PTrees, the model comprehensively understands the complex relationships within the data, thereby enhancing its predictive capabilities.
\subsubsection{SHapley Additive exPlanations (SHAP)}
To enhance model interpretability and elucidate feature importance, SHAP \cite{biecek2021explanatory} was integrated into the framework.
SHAP values  provide a unified measure of feature importance, enabling us to understand the contribution of each feature to the model’s predictions. This enhances the transparency and trustworthiness of our model's outputs.
Rooted in cooperative game theory, SHAP values offer a method to attribute the difference between the prediction for a specific instance and the average prediction to individual features. SHAP values adhere to local accuracy, missingness, and consistency, ensuring reliable and interpretable explanations.
Mathematically, for a model \( f \) and an instance \( x \), the SHAP value \( \phi_i \) for feature \( i \) is calculated as:

\[
\phi_i = \sum_{S \subseteq N \setminus \{i\}} \frac{|S|! (|N| - |S| - 1)!}{|N|!} \left[ f(S \cup \{i\}) - f(S) \right]
\]

where:
\begin{itemize}
    \item \( N \) is the set of all features.
    \item \( S \) is a subset of \( N \) that excludes feature \( i \).
    \item \( f(S) \) is the prediction for the instance with only the features in \( S \).
\end{itemize}

This formula calculates the average marginal contribution of feature \( i \) across all possible feature subsets, ensuring fair and comprehensive feature importance attribution.

To facilitate SHAP analysis, predictions from the ensemble of PTrees were encapsulated in a wrapper compatible with the SHAP framework. A background dataset was generated using k-means clustering on the training data to provide a reference point for SHAP value calculations. SHAP values were computed for a subset of the test data using the Kernel Explainer to balance computational efficiency and accuracy 
\begin{algorithm}
\caption{Create Tree from Data}\label{alg:create_tree}
\begin{algorithmic}[1]
\Require $father\_node$: Node, $data$: DataFrame, $variable\_order$: List, $level$: Integer, $pruning\_threshold$: Float
\Ensure Root node of the decision tree ($father\_node$)

\Function{create\_tree\_from\_data}{$father\_node$, $data$, $variable\_order$, $level$, $pruning\_threshold$}
\State $current\_variable \gets variable\_order[0]$
\State $current\_data \gets data[current\_variable]$
\State $class\_nodes \gets$ empty list
\For{$val$ in unique values of $current\_data$}
    \State Create class node with ID, level, statements, and no children
    \State Append class node to $class\_nodes$
\EndFor

\State $total\_samples \gets$ total number of samples in $current\_data$

\For{each $class\_node$ and $val$ in $class\_nodes$}
    \State $is\_root \gets$ Check if $father\_node$ is root
    \State $val\_count \gets$ Count of $val$ in $current\_data$
    \If {$is\_root$}
    \State Calculate transition probability based on occurrences
    \If {$transition\_prob \geq pruning\_threshold$}
    \State Insert transition probability into $father\_node$
    \EndIf
    \Else 
    \State Calculate transition probability based on parent's state
    \If {$transition\_prob \geq pruning\_threshold$}
    \State Insert transition probability into $father\_node$
    \EndIf
    \EndIf
\EndFor

\For{each $class\_node$ and $val$ in $class\_nodes$}
    \State Get next data for $val$
    \State Get next variable order
    \If{next variable order is not empty}
    \State Recursively call $create\_tree\_from\_data$ 
    \EndIf
\EndFor

\State \Return $father\_node$
\EndFunction
\end{algorithmic}
\end{algorithm}

\begin{algorithm}
\caption{Conditional Probability Calculation}
\label{alg:conditional_probability}
\begin{algorithmic}[1]
\Function{conditionalProbability}{$self$,$\textit{input\_condition}$}
\If{$\textit{input\_condition}$ is empty}
\State \Return $0.0$
\EndIf

\State $\textit{cut\_disease} \gets \text{self.prop('target\_variable')}$
\State $\textit{combined\_cut} \gets \text{None}$

\ForAll{$(\textit{var}, \textit{val})$ in $\textit{input\_condition}$}
\State $\textit{cut} \gets \text{self.prop(var + ' = ' + val)}$
\If{$\textit{combined\_cut}$ is None}
\State $\textit{combined\_cut} \gets \textit{cut}$
\Else
\State $\textit{combined\_cut} \gets \textit{combined\_cut} \land \textit{cut}$
\EndIf
\EndFor
\State $\textit{disease\_see} \gets \text{self.see}(\textit{combined\_cut})$

\State $\textit{probability} \gets \textit{disease\_see.prob}(\textit{cut\_disease})$

\State \Return $\textit{probability}$
\EndFunction
\end{algorithmic}
\end{algorithm}

\section{Case Studies} \label{casestudies}
The proposed framework addresses several limitations of traditional prediction models by offering a multifaceted approach for clinicians. It facilitates the identification of causal relationships between variables, enables predictive modeling, and supports the exploration of potential interventions and counterfactual scenarios. This combination provides clinicians with a more comprehensive understanding of the data while acknowledging the inherent challenges of causal analysis.

We evaluated the framework using multiple real-world healthcare datasets to assess its applicability and generalisability across diverse clinical contexts. The first dataset was MIMIC-IV, a collection of electronic health records from critical care settings, where the objective was to predict the length of stay in the Intensive Care Unit (ICU). The second dataset was the Framingham Heart Study, which focuses on cardiovascular disease (CVD) and its risk factors, trends over time, and familial patterns. Finally, the Diabetes dataset from BRFSS-2015 was used to analyse risk factors and predict the likelihood of diabetes onset. These case studies were selected to evaluate the framework's utility in diverse scenarios and assess its ability to handle distinct clinical challenges. 

\subsection{Length of stay in ICU case study}
The intensive care unit (ICU) stands as a vital line of defense for critically ill patients, offering specialised care to prevent deterioration from severe illness or injury \cite{marshall2017intensive}, \cite{weil2011intensive}. However, the ever-increasing demand for ICU beds threatens this critical service. The imbalance between ICU capacity and patient needs has significant consequences for patient outcomes, public health, and even socio-economic factors \cite{robinson1966prediction}, \cite{stone2022systematic}. Therefore, optimising ICU resource allocation and planning for future needs necessitates interpretable models that facilitate counterfactual analysis for informed decision-making, ultimately ensuring optimal care for critically ill patients.

\subsubsection{MIMIC-IV} 
In this study, we use the MIMIC-IV version 2.2 database \cite{Johnson2023MIMIC}, which includes patients admitted to the BETH Israel Deaconess Medical Center during the period 2008–2019. The data contains multiple dimensions, from administrative data to laboratory results and diagnoses.We employed preprocessing techniques as described in \cite{mdpi} to ensure consistency and comparability with existing literature. The cohort included all patients with at least one ICU visit. However, certain subsets of patients were excluded: those who died during their ICU stay, those who returned to the ICU within 48 hours of discharge, those with an LOS greater than 21 days, and those with an LOS of less than one day.

The exclusion of patients who returned to the ICU within 48 hours was motivated by the focus of this analysis on understanding factors influencing the initial ICU stay and its length. Rapid readmissions often reflect distinct cases with underlying complexities such as incomplete recovery or premature discharge, which could introduce confounding factors. Similarly, patients with extremely long LOS (greater than 21 days) were excluded to avoid the influence of outliers, which could disproportionately impact model performance. Patients with an LOS of less than one day were excluded because the data collected during the first 24 hours was used for modeling, making such cases incomplete for analysis. These exclusions ensure that the cohort is representative of the broader ICU patient population, allowing for more generalisable findings. Future work could investigate the effects of these exclusions on model performance by reintroducing these subsets for a comparative analysis.

To transform the length-of-stay task into a classification problem, we categorised LOS into clinically meaningful groups: short stays (1–4 days) and long stays (greater than 4 days). This categorisation was guided by the 75th percentile of LOS distribution (Q3 = 4.0) in the dataset, as described in \cite{mdpi}, and reflects thresholds commonly used in critical care practice.

\subsection{Heart Disease case study}
Despite significant advancements in healthcare, CHD remains a leading cause of global mortality, accounting for 17.9 million deaths in 2019 as reported by the World Health Organization (WHO) \cite{who_cvds_2021}. While accurate prediction of future risk is undeniably crucial, medical experts increasingly recognise the limitations of solely relying on such prognostic models. To optimise patient care, a deeper understanding of the factors influencing individual susceptibility to CHD is paramount. This necessitates the development of intelligent systems that can not only predict future risk but also explore the potential impact of interventions and counterfactual analysis.
\subsubsection{Framingham Data} 
In this study, we use the Framingham heart disease dataset includes over 4238 records and 15 attributes \cite{mahmood2014framingham}. The goal of the dataset is to predict whether the patient has 10-year risk of future CHD. The initial preprocessing steps involved converting the numerical variables in the dataset into categorical variables. Given that the variables pertain to health-related data, specific ranges were meticulously considered during this conversion process. Numerical data representing health metrics such as blood pressure, cholesterol levels, or body mass index were categorised into clinically relevant ranges indicative of different health conditions or risk levels. By transforming numerical data into categorical form based on meaningful health-related ranges, the dataset became better suited for subsequent analysis and interpretation within the context of healthcare applications.

\subsection{Diabetes case study}
Despite the existence of preventative measures, diabetes remains a significant global health burden \cite{ali2017global}. Characterised spectrum of devastating complications, it necessitates a multifaceted approach that transcends traditional risk prediction. While accurate future risk prediction remains valuable for preventative strategies, a deeper understanding of modifiable factors influencing individual susceptibility is paramount, especially considering the potential for early intervention to reduce diabetes-related mortality \cite{an2024early}. This necessitates the development of robust computational models capable of not only predicting future risk but also exploring the potential impact of various interventions through counterfactual analysis. Such models could empower clinicians by enabling the exploration of "what-if" scenarios: investigating how a patient's risk profile might change with different lifestyle modifications or therapeutic interventions, ultimately leading to tailored preventative strategies and optimised patient care. 

\subsubsection{Diabetes Data} 
Data was obtained from the Behavioral Risk Factor Surveillance System (BRFSS), which is the primary system of health-related telephone surveys that collect state data on risk behaviours, chronic health conditions, and use of preventative treatments amongst U.S. residents \cite{centers}. The survey started in 1984 and currently performs over 400,000 adult interviews each year, making it the world's largest continuously conducted health survey system. This survey data provides a dataset that could be used to analyse and forecast diabetes risk variables. We utilised the BRFSS-2015 dataset, which included 253,680 health assessments. 

\section{Evaluation and discussion of the results}\label{results}
The evaluation process begins in Section \ref{sensitivityanalysis} with an analysis of the varying outcomes derived from the sensitivity analysis. In Section \ref{subsec:prediction}, we assess the predictive performance of the PCF model, comparing it against a range of benchmark models, including both interpretable and non-interpretable methods, across all three datasets. Additionally, this section explores model interpretability using SHAP. Section \ref{subsec:interventions} then examines the effects of potential interventions through interventional analysis. Lastly, Section \ref{subsec:counterfactual} investigates counterfactual analysis to provide further insights.

\subsection{Interpretation of Sensitivity Analysis}
\label{sensitivityanalysis}
\begin{figure*}[ht]
\centering
 \includegraphics[width=14cm]{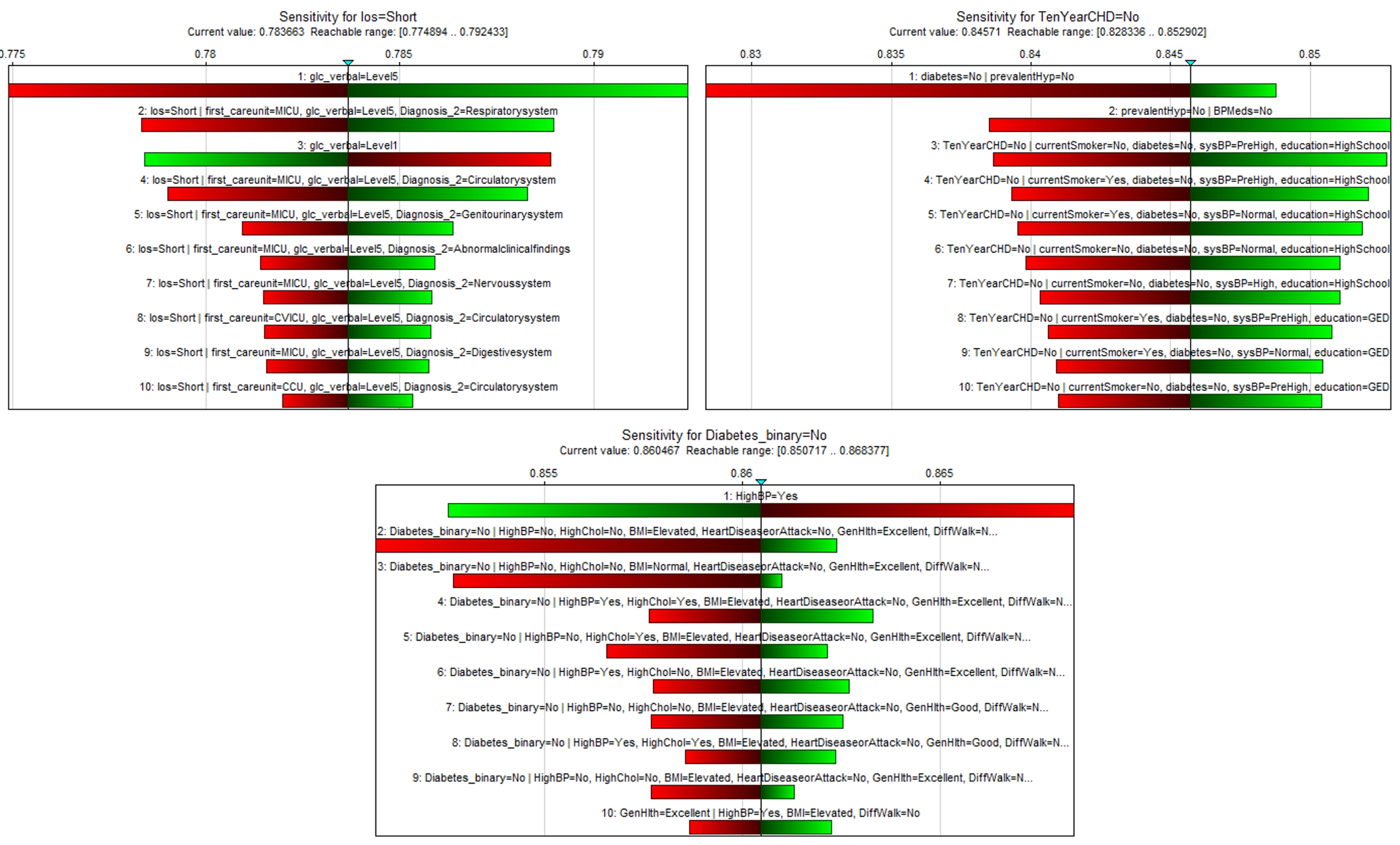}
    \caption{Sensitivity Analysis for LOS, Diabetes, and Framingham datasets.}
    \label{fig:sensitivity_analysis}
    
\end{figure*}

Sensitivity analysis is a crucial step in our framework to understand the influence of various parameters on the target variable.
The diverse outcomes from our sensitivity analysis provide valuable insights into the multifaceted factors influencing LOS, CHD, and Diabetes, as depicted in Figure \ref{fig:sensitivity_analysis}. The color of the bars indicates the direction of change in the target state, with red representing a negative impact and green representing a positive impact.
For LOS, factors such as first care unit admission, patient's verbal communication ability, and specific diagnoses (e.g., Respiratory system, Circulatory system) show high sensitivity, indicating their collective substantial impact on LOS. In the context of CHD, the absence of diabetes and hypertension was found to significantly reduce the risk. Additionally, other significant factors include being a non-smoker with normal systolic blood pressure and specific education levels. These findings highlight the combined effect of lifestyle and socio-economic factors on CHD risk. For Diabetes, the sensitivity analysis demonstrates the complex interplay between hypertension, cholesterol levels, BMI, and other health indicators. The figure reveals that individuals with
hypertension have the highest sensitivity value, indicating that high blood pressure significantly increases the risk of developing diabetes. Additionally, other influential factors include high cholesterol, elevated BMI, and the presence of heart disease. These findings underscore the complex interplay of multiple health conditions in determining diabetes risk, highlighting the necessity of addressing various health parameters simultaneously to effectively manage and prevent diabetes.
\subsection{Prediction}
\label{subsec:prediction}
This section evaluates the predictive capabilities of PCF by applying it to three clinical datasets.  We conduct a comprehensive assessment of its performance by juxtaposing its outcomes against those attained by established benchmark methodologies, comprising Logistic Regression (LR) \cite{lavalley2008logistic}, Decision Tree (DT) \cite{suthaharan2016decision}, Random Forest (RF) \cite{rigatti2017random}, Support Vector Machine (SVM) \cite{vapnik1995nature}, K-Nearest Neighbors (KNN)\cite{nn}, \cite{Mucherino2009}, Gradient Boosting (GB) \cite{FREUND1997119}, \cite{gradientboosting}, eXtreme Gradient Boosting (XGB) \cite{xgb}, and Adaptive Boosting (Adaboost) \cite{freund1996experiments}. Noteworthy among these benchmarks are LR, DT, and KNN, esteemed for their interpretability, which facilitates the elucidation of their decision-making mechanisms. Additionally, given the ensemble nature of our approach involving PTrees, a comparison with ensemble techniques such as GB, XGB, Adaboost and RF is warranted.

The dataset undergoes stratification-based partitioning into training and testing subsets to ensure their representativeness. Subsequently, the performance of each model is assessed utilising diverse metrics such as accuracy, specificity, sensitivity, and the Area Under the Receiver Operating Characteristic Curve (AUC-ROC). Predictions are made based on a default threshold, with the potential for adjustment in scenarios characterised by resource constraints, thereby prioritising cases of utmost urgency.

The performance metrics for each dataset are presented in Table \ref{tab:mimic_results}, \ref{tab:fram_results} and \ref{tab:diabetes_results}. It is well-documented that class imbalance within datasets can significantly impact the evaluation of machine learning algorithms. To mitigate this potential bias and ensure a fair comparison across all models, various techniques for handling class imbalance were employed. In the case of the MIMIC-IV and Framingham heart datasets, the SMOTE (Synthetic Minority Over-Sampling Technique) \cite{chawla2002smote} oversampling technique yielded superior results. Conversely, ADASYN (Adaptive Synthetic Minority Oversampling Technique) \cite{he2008adasyn} demonstrated the best performance when applied to the diabetes dataset. This finding suggests that the most effective class imbalance handling technique may vary depending on the specific characteristics of the data and the machine learning models being evaluated. 

\begin{table}[ht]
\scriptsize
\caption{Results using SMOTE for MIMIC-IV}
\label{tab:mimic_results}
\begin{adjustbox}{max width=\linewidth}
\begin{tabularx}{\linewidth}{lXXXX}
\hline
\textbf{Algorithm} & \textbf{Accuracy} & \textbf{Specificity} & \textbf{Sensitivity} & \textbf{AUC-ROC} \\
\hline
\multicolumn{5}{l}{\textbf{Ensemble Algorithms}} \\
\hline
GB & 73.71 & 77.96 & 57.91 & 67.94 \\
XGB & 73.92 & 79.14 & 54.54 & 66.84 \\
Adaboost & 75.07 & 79.78 & 57.57 & 68.67 \\
RF & 75.64 & 83.22 & 47.47 & 65.35 \\
\textbf{PCF} & \textbf{73.21} & \textbf{77.69} & \textbf{56.56} & \textbf{67.13} \\
\hline
\multicolumn{5}{l}{\textbf{Other Algorithms}} \\
\hline
SVM & 73.07 & 76.42 & 60.60 & 68.51 \\
KNN & 71.14 & 77.87 & 46.12 & 62.00 \\
\hline
\multicolumn{5}{l}{\textbf{Interpretable Algorithms}} \\
\hline
DT & 79.42 & 88.84 & 44.44 & 66.64 \\
LR & 70.14 & 73.70 & 56.90 & 65.30 \\
PTree & 80.29 & 91.11 & 40.06 & 65.59 \\
\hline
\end{tabularx}
\end{adjustbox}
\end{table}

\begin{table}[ht]
\scriptsize
\caption{Results using SMOTE for Framingham heart data}
\label{tab:fram_results}
\begin{adjustbox}{max width=\linewidth}
\begin{tabularx}{\linewidth}{lXXXX}
\hline
\textbf{Algorithm} & \textbf{Accuracy} & \textbf{Specificity} & \textbf{Sensitivity} & \textbf{AUC-ROC} \\
\hline
\multicolumn{5}{l}{\textbf{Ensemble Algorithms}} \\
\hline
GB & 63.08 & 64.81 & 53.48 & 59.15 \\
XGB & 63.91 & 68.01 & 41.08 & 54.54 \\
Adaboost & 66.03 & 69.26 & 48.06 & 58.66 \\
RF & 67.09 & 72.73 & 35.65 & 54.19 \\
\textbf{PCF} & \textbf{66.98} & \textbf{69.68} & \textbf{51.93} & \textbf{60.80} \\
\hline
\multicolumn{5}{l}{\textbf{Other Algorithms}} \\

\hline
SVM & 69.22 & 74.26 & 41.08 & 57.67 \\
KNN & 76.76 & 87.62 & 16.27 & 51.95 \\
\hline
\multicolumn{5}{l}{\textbf{Interpretable Algorithms}} \\
\hline
DT & 64.03 & 66.06 & 52.71 & 59.38 \\
LR & 61.55 & 63.14 & 52.71 & 57.92 \\
PTree & 65.57 & 70.23 & 39.53 & 54.88 \\
\hline
\end{tabularx}
\end{adjustbox}
\end{table}

\begin{table}[ht]
\scriptsize
\caption{Results using ADASYN for diabetes data}
\label{tab:diabetes_results}

\begin{adjustbox}{max width=\linewidth}
\begin{tabularx}{\linewidth}{lXXXX}
\hline
\textbf{Algorithm} & \textbf{Accuracy} & \textbf{Specificity} & \textbf{Sensitivity} & \textbf{AUC-ROC} \\
\hline
\multicolumn{5}{l}{\textbf{Ensemble Algorithms}} \\
\hline
GB & 70.78 & 70.02 & 75.66 & 72.84 \\
XGB & 69.71 & 71.34 & 59.25 & 65.30 \\
Adaboost & 71.35 & 71.09 & 73.01 & 72.05 \\
RF & 73.42 & 77.45 & 47.61 & 62.53 \\
\textbf{PCF} & \textbf{73.64} & \textbf{73.41} & \textbf{75.13} & \textbf{74.27} \\
\hline
\multicolumn{5}{l}{\textbf{Other Algorithms}} \\
\hline
SVM & 69.71 & 68.62 & 76.71 & 72.67 \\
KNN & 79.00 & 85.86 & 35.26 & 60.56 \\
\hline
\multicolumn{5}{l}{\textbf{Interpretable Algorithms}} \\
\hline
DT & 62.92 & 60.19 & 80.42 & 70.31 \\
LR & 70.71 & 70.27 & 73.54 & 71.90 \\
PTree & 72.36 & 71.12 & 52.82 & 61.97 \\
\hline
\end{tabularx}
\end{adjustbox}
\end{table}

Tables \ref{tab:mimic_results}, \ref{tab:fram_results}, and \ref{tab:diabetes_results} present the performance evaluation of the PCF framework compared to benchmark methodologies across the three datasets. Notably, PCF achieves results that are largely comparable to established ensemble-based and interpretable models, balancing specificity and sensitivity while maintaining competitive predictive accuracy.

On the MIMIC-IV dataset, PCF achieves an accuracy of 73.21\% and an AUC-ROC of 67.13\%, performing similarly to ensemble-based methods such as Gradient Boosting (73.71\% accuracy) and Adaboost (75.07\% accuracy). Although DT achieves a higher accuracy of 79.42\%, PCF demonstrates better sensitivity (56.56\%) than simpler models like KNN, which achieves only 46.12\%. Among ensemble methods, PCF effectively balances specificity and sensitivity, avoiding extremes like RF, which prioritises specificity at the expense of sensitivity.

On the Framingham dataset, PCF achieves an AUC-ROC of 60.80\%, outperforming most ensemble methods while maintaining competitive accuracy at 66.98\%, compared to RF (67.09\%) and Adaboost (66.03\%). Although KNN achieves the highest accuracy at 76.76\%, its significantly lower sensitivity (16.27\%) highlights its limitations in handling balanced classification scenarios. PCF’s balanced trade-off between specificity and sensitivity makes it particularly well-suited for datasets requiring nuanced predictions.

On the diabetes dataset, PCF achieves the highest AUC-ROC among all models (74.27\%) and an accuracy of 73.64\%, comparable to ensemble methods such as RF (73.42\%) and higher than DT (62.92\%). While KNN achieves the highest accuracy (79.00\%), its specificity (85.86\%) comes at the cost of sensitivity (35.26\%). PCF balances both metrics effectively, achieving 73.41\% specificity and 75.13\% sensitivity, demonstrating its robustness across diverse datasets.

Overall, while PCF does not consistently outperform simpler models such as DT or KNN in terms of accuracy, it offers balanced performance across key metrics and demonstrates robustness across datasets. Its ability to maintain competitive predictive performance while also uncovering causal relationships and supporting intervention modeling sets it apart from purely predictive methodologies.

\subsubsection{Interpretability with SHAP}
\begin{figure*}[ht]
\centering
 \includegraphics[width=14cm]{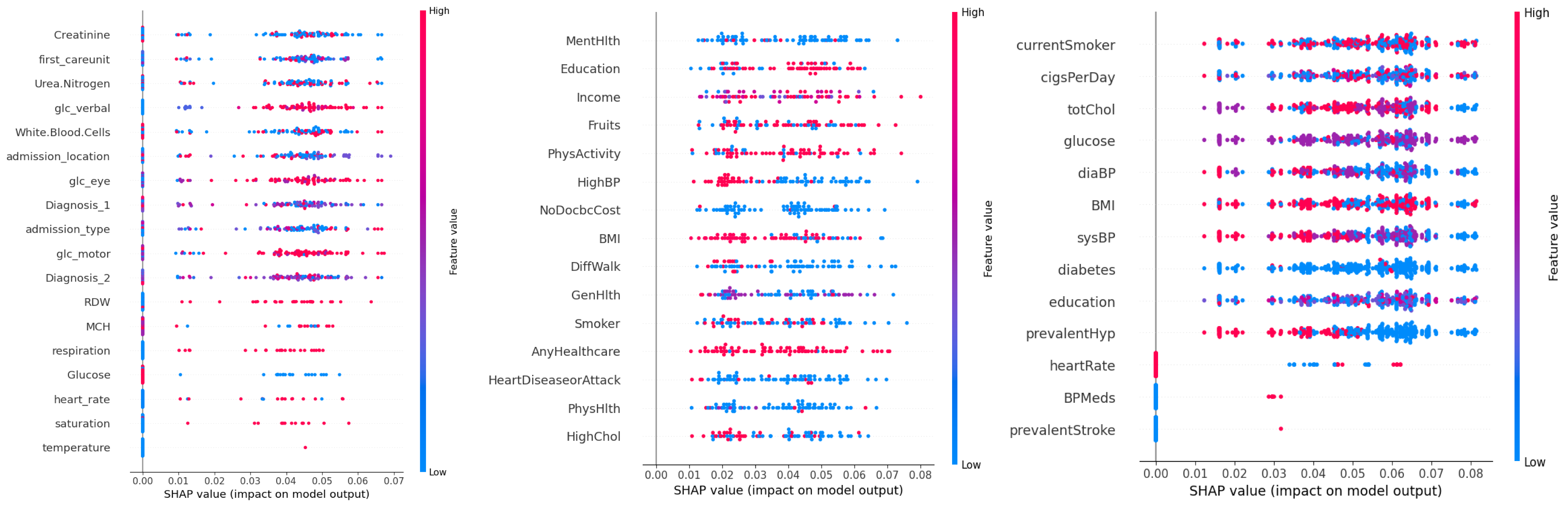}
    \caption{SHAP plot showing feature impacts on predictions for LOS, CHD, and Diabetes.}
    \label{fig:SHAP}
    
\end{figure*}

The SHAP plot, shown in Figure \ref{fig:SHAP}, interprets the influence of each feature on predictions for LOS, Coronary Heart Disease (CHD), and Diabetes. For LOS, features such as creatinine, first care unit, and urea nitrogen have high SHAP values, indicating their strong influence on prolonged ICU stays. The Glasgow Coma Scale (glc\_verbal) score and elevated white blood cells, along with specific diagnoses (e.g., respiratory and circulatory system issues), also significantly impact LOS predictions.
In CHD predictions, critical features include current smoking status, systolic blood pressure (sysBP), and glucose levels, which are known risk factors for heart disease. High cholesterol (totChol), the presence of diabetes, socio-economic factors, and lifestyle choices such as education level and physical activity further influence CHD risk.
For Diabetes, key contributors include high blood pressure (HighBP), elevated BMI, and high cholesterol levels, which are essential components of metabolic syndrome. General health status (GenHlth) and difficulty walking (DiffWalk) also play significant roles, along with the presence of heart disease and levels of physical activity.
The SHAP plots reveal that LOS is heavily influenced by clinical indicators related to critical health conditions and specific ICU units. CHD risk is predominantly affected by cardiovascular risk factors, lifestyle choices, and socio-economic factors, while Diabetes risk is determined by metabolic health markers, overall health, and physical activity. These insights validate the results from sensitivity analysis and provide a detailed understanding of feature contributions, helping identify key areas for intervention and improve clinical decision-making processes by emphasising the most impactful factors for each health condition.
By combining sensitivity analysis for understanding variable influence within the CBN and SHAP values for detailed prediction explanations, our framework ensures robust causal inference and clear, actionable insights for clinical decision-making.

\subsection{Intervention} \label{subsec:interventions}
\begin{figure*}[ht]
\centering
 \includegraphics[width=14cm]{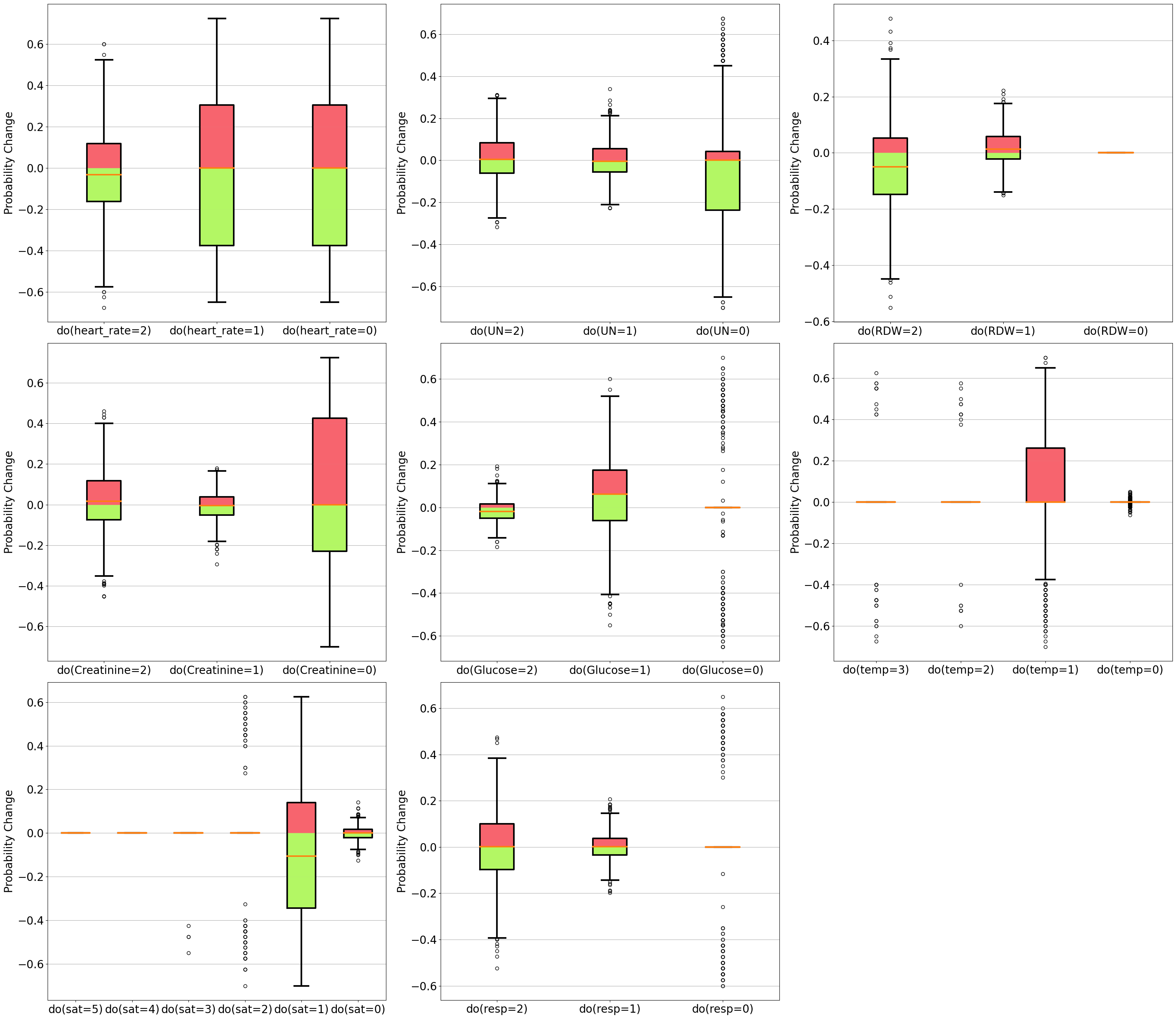}
    \caption{Probability change of los given interventions on heart\_rate, Urea\_Nitrogen (UN), RDW, Creatinine, Glucose, temperature (temp), saturation (sat), and respiration rate (resp)}
    \label{fig:los_interven}
    
\end{figure*}

In the context of PCF, intervention involves the strategic modification of transition probabilities to ensure a specific event occurs with certainty (probability of 1). This approach allows for the exploration of conditional probabilities represented as $P(A | do(B))$, indicating the probability of event $A$ occurring given that event $B$ is enforced. Unlike in CBNs, interventions in PCF are more general and do not require unique value assignments to manipulated random variables. Instead, the impact of an intervention depends on the critical set, allowing for different values for the same random variables in various branches of the tree. Moreover, it can manipulate different random variables in each branch, functioning as a context-dependent "recipe" to achieve the desired outcome.
\paragraph{\textbf{MIMIC-IV:}}
To elucidate the impact of key physiological parameters on ICU length of stay (LOS), an interventional analysis was conducted. Eight critical factors were examined to assess the PCF model's ability to replicate established causal relationships. The primary objective was to assess PCFs ability to replicate established causal relationships between these parameters and LOS.Box plots (Figure \ref{fig:los_interven}) were used to visualise the distribution of los across different intervention groups. In these plots, red signifies an increased probability of los exceeding 4 days (los = 1), while green signifies a decrease.

Existing literature establishes a link between bradycardia (heart rate $\leq$ 60 beats per minute) and extended ICU stays due to underlying medical conditions requiring further investigation or treatment \cite{heartrate}. To explore this relationship within our model, interventional analysis was conducted on heart rate. Simulating bradycardia ($do(heart\_rate=0)$) significantly increased the likelihood of extended ICU stays ($los=1$), consistent with established medical knowledge. However, the relationship between heart rate and length of stay is more complex. Similar trends were observed for heart rates above 60 bpm ($do(heart\_rate=1)$), though to a lesser extent, and a slight decrease in $los=1$ was noted for even higher heart rates ($do(heart\_rate=2)$).This highlights the need to consider multiple physiological and clinical variables beyond heart rate. 

In literature, it is established that low levels of urea nitrogen (UN) are not typically concerning, often associated with low protein intake \cite{urea}. Similar findings are observed in our model, where manipulating UN levels to be low ($do(Urea\_Nitrogen = 0)$) results in a decrease in the proportion of patients with extended ICU stays ($los=1$). However, as the intervention values increase ($do(Urea\_Nitrogen = 1)$ and $do(Urea\_Nitrogen = 2)$), a trend emerges indicating a potential rise in the probability of $los=1$. While this increase is subtle, it is visually detectable by the red color in the box plots.

Literature indicates that elevated Red Cell Distribution Width (RDW) is closely associated with increased risk of cardiovascular morbidity and mortality in patients with previous myocardial infarction, potentially leading to prolonged hospital stays \cite{tonelli2008relation}. Our model's interventional analysis, where RDW is manipulated to be high ($do(RDW = 2)$), shows a corresponding increase in the percentage of patients with extended ICU stays ($los=1$), consistent with existing literature.

Lower levels of creatinine are often related to muscle loss and severe liver disease. Patients experiencing significant muscle mass loss in the first week of ICU admission are at higher risk of extended stays \cite{de2023good}.This aligns with our findings, where intervening to set low creatinine levels ($do(Creatinine = 0)$) notably increases the likelihood of extended ICU stays ($los=1$).

High respiration rate, indicative of tachypnea in adults, is characterised by a respiratory rate exceeding 20 breaths per minute and often requires further assessment, leading to prolonged hospital stays \cite{Puskarich2017AssociationBP}. In our model, intervening to elevate the respiration rate ($do(respiration = 2)$) resulted in an increased probability of extended ICU stays ($los=1$).

Fever is a common issue in ICU patients and often necessitates diagnostic tests and procedures, which significantly prolongs the stay \cite{cunha1996fever}. Consistent with this, our model shows that intervening to indicate mild fever ($do(temperature = 1)$) also leads to an elevated probability of extended ICU stays ($los=1$).

Inadequate oxygen saturation ($do(saturation = 1)$) has a varied effect on $los=1$, while other saturation levels show no discernible impact. Manipulating glucose levels reveals an inverse response, suggesting these variables indirectly influence LOS through intermediary factors rather than exerting a direct effect.

\begin{figure*}[ht]
\centering
 \includegraphics[width=14cm]{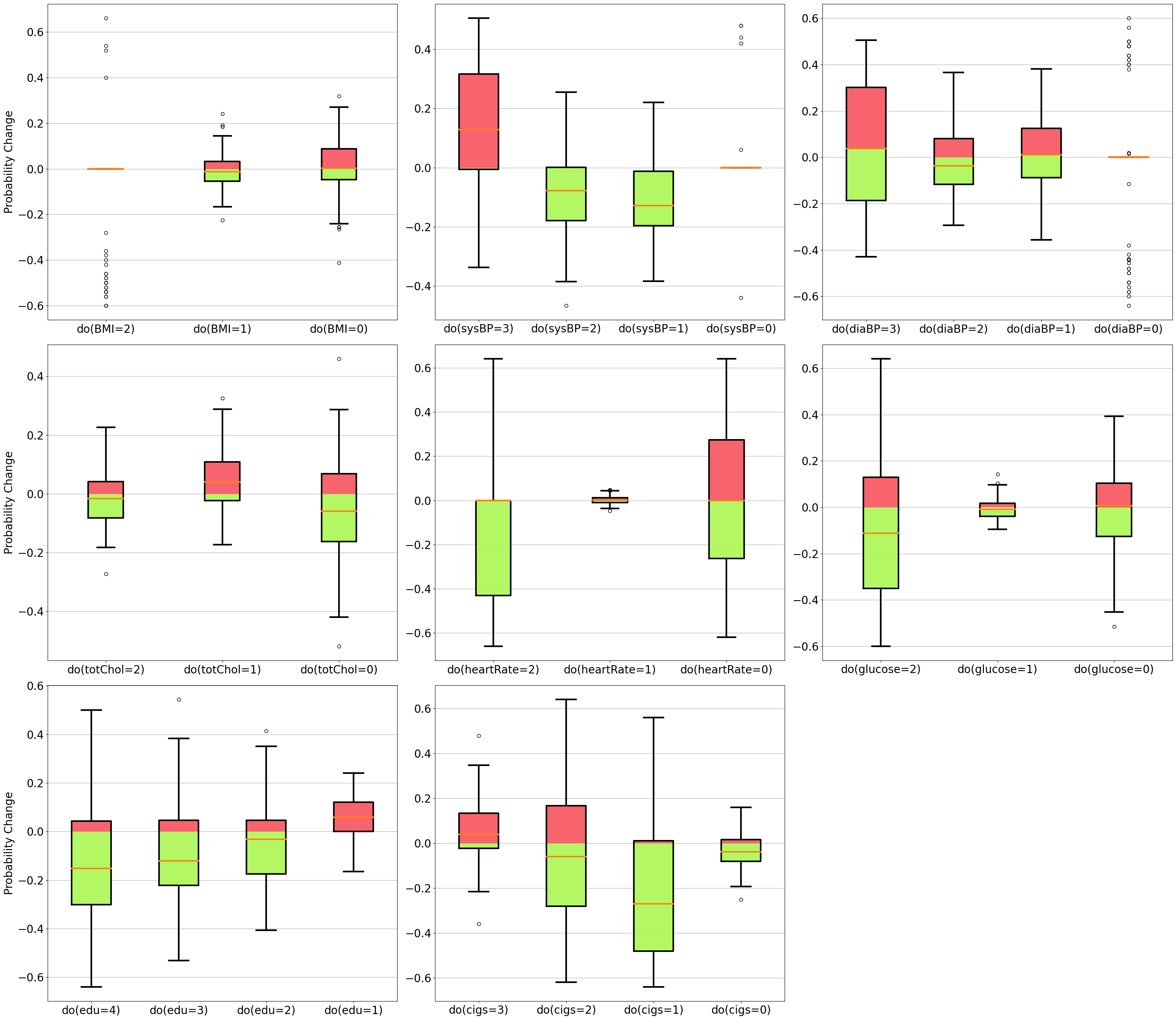}
    \caption{Probability change of TenYearCHD given interventions on sysBP, diaBP, totChol, BMI, education, glucose, heartRate and cigsPerDay.}
    \label{fig:fram_interven}
    
\end{figure*}

\paragraph{\textbf{Framingham Heart Data:}}
This section explores the impact of various health factors on CHD risk through intervention analysis, aiming to assess the PCF's ability to replicate established causal relationships between these parameters and CHD. As illustrated in \ref{fig:fram_interven}, our findings underscore significant alterations in $P(TenYearCHD= 1)$ across different interventions, shedding light on the intricate interplay between these health factors and CHD risk.

Existing literature highlights both systolic and diastolic hypertension as independent risk factors for adverse cardiovascular events \cite{BP}. Our analysis corroborates this, demonstrating that interventions on systolic blood pressure, such as $do(sysBP=3)$, result in an increased probability of $CHD = 1$. Similarly, interventions on diastolic blood pressure ($do(diaBP=1)$, $do(diaBP=2)$ and $do(diaBP=3)$) also heighten the likelihood of $CHD = 1$ , reinforcing the significant impact of blood pressure levels on cardiovascular health.

Additionally, glucose metabolism plays a critical role in cardiovascular health, as deviations from normal glucose levels can lead to adverse outcomes \cite{glucose}. Our model confirms this relationship, showing that interventions altering glucose levels, such as $do(glucose=0)$ and $do(glucose=2)$ , significantly increase the probability of $CHD = 1$. These findings underscore the critical role that glucose regulation plays in cardiovascular risk management.

Raised total cholesterol levels are well-documented as a significant risk factor for coronary heart disease (CHD) \cite{peters2016total}. In our model, interventions on total cholesterol ($do(totChol=1)$ i.e levels $\geq$ 200) were found to increase the probability of $CHD = 1$, reinforcing the established link between elevated cholesterol and heightened CHD risk.

Literature highlights that asymptomatic bradycardia may influence heart disease risk due to underlying autonomic or cardiovascular issues \cite{dharod2016association}. Our intervention analysis, which simulates bradycardia through interventions on heart rate ($do(heartRate=0)$), reveals a marked increase in the probability of $CHD = 1$.

Smoking has been highlighted as a leading risk factor for heart disease \cite{gallucci2020cardiovascular}. Our model's interventions demonstrate that smoking 6-10 cigarettes per day ($do(cigsPerDay =2)$) and more than 11 cigarettes per day ($do(cigsPerDay =3)$) significantly increase the probability of $CHD = 1$. These findings underscore the substantial impact of smoking on coronary heart disease risk.

Research indicates that higher education levels can lead to substantial health benefits \cite{education}. Our model corroborates these findings, demonstrating that higher levels of education significantly decrease the probability of $CHD = 1$.

The relationship between BMI and CHD is often characterised as inconsistent and complex \cite{bmi}. Our findings support this observation, as BMI interventions did not produce interpretable results. This ambiguity may be attributed to the intricate interplay of metabolic factors that extend beyond body mass alone, suggesting that BMI might not be a straightforward predictor of CHD risk. The lack of a clear relationship underscores the need for a more nuanced understanding of how metabolic factors contribute to CHD.

\begin{figure*}[ht]
\centering
 \includegraphics[width=14cm]{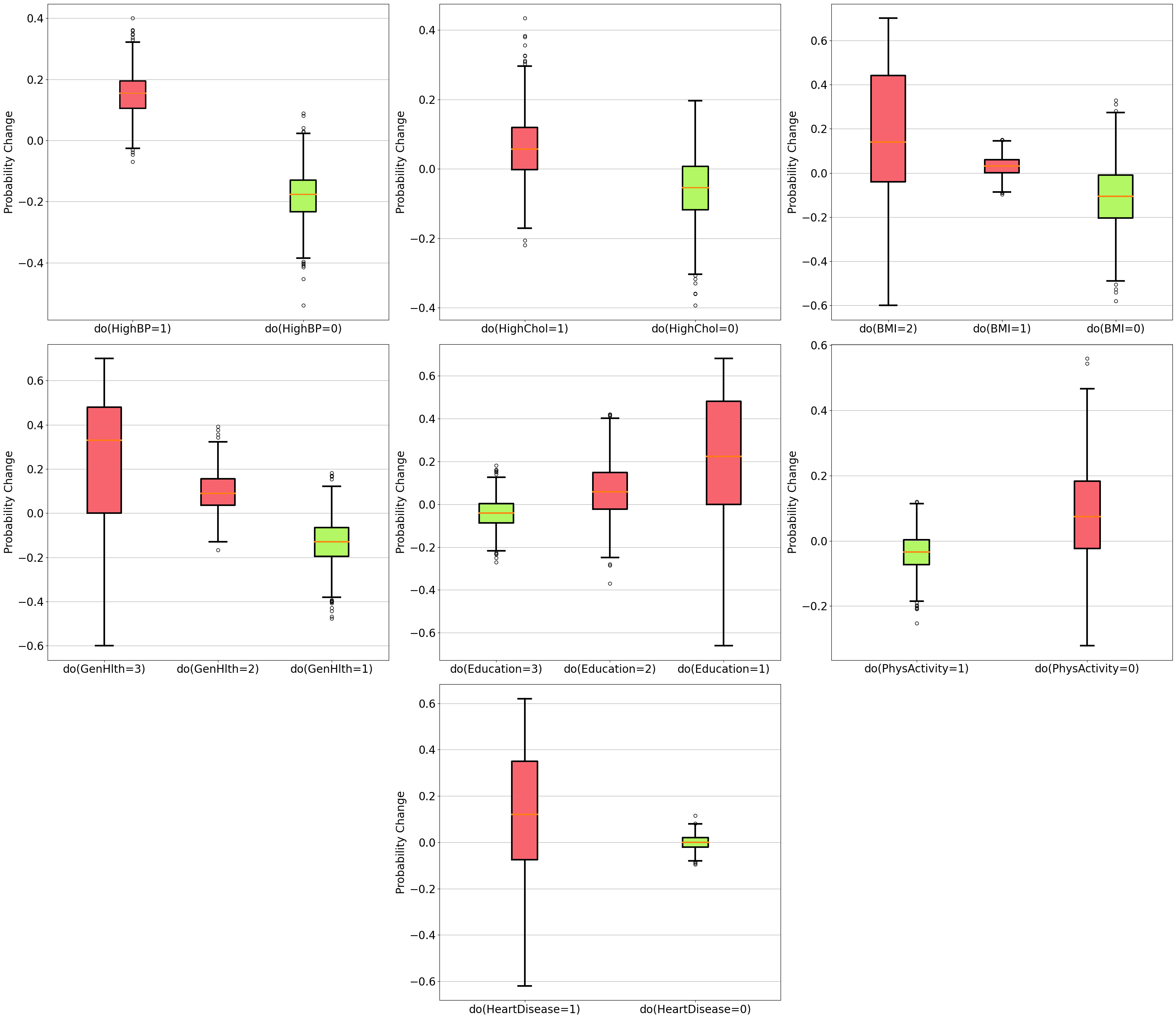}
    \caption{Probability change of Diabetes given interventions on sysBP, diaBP, totChol, BMI, education, glucose, heartRate and cigsPerDay.}
    \label{fig:diab_interven}
    
\end{figure*}

\paragraph{\textbf{Diabetes:}}
This section explores the influence of various health factors on the likelihood of developing diabetes through intervention analysis. 
As depicted in Figure \ref{fig:diab_interven}, illustrates the significant changes in diabetes risk ($P(Diabetes = 1)$) following interventions on the selected variables.

Hypertension and hyperlipidemia are well-established predictors of diabetes risk, as highlighted in existing literature \cite{wei2011blood, cholesterol}. Our analysis confirms this relationship, showing that high blood pressure ($do(HighBP=1)$) increases diabetes probability, while its absence ($do(HighBP=0)$) decreases the risk. Similarly, elevated cholesterol levels ($do(HighChol=1)$) are linked to a higher likelihood of diabetes, whereas normal cholesterol levels ($do(HighChol=0)$) reduce the risk.

Body Mass Index (BMI) is another significant risk factor for diabetes \cite{escBMI2020}. Our findings indicate that maintaining a normal BMI ($do(BMI=0)$) lowers the probability of diabetes, while a BMI of 40 or more ($do(BMI=2)$) substantially raises this probability. This suggests that keeping a BMI between 0-24 mitigates diabetes risk, whereas higher BMI levels considerably elevate it.

Maintaining a healthy lifestyle is crucial for diabetes prevention \cite{phys}. Our analysis demonstrates that individuals with excellent general health ($do(GenHlth=1)$) have a lower risk of developing diabetes compared to those in good ($do(GenHlth=2)$) or poor health ($do(GenHlth=3)$). Additionally, regular physical activity ($do(PhysActivity=1)$) significantly reduces diabetes risk compared to a sedentary lifestyle ($do(PhysActivity=0)$).

Education also plays a positive role in diabetes management and complication prevention \cite{sil2020study}. Our results indicate that individuals with limited education ($do(education=1)$) face a higher diabetes risk, while those with some secondary education ($do(education=2)$) show mixed outcomes. Notably, higher education levels ($do(education=3)$) are significantly associated with reduced diabetes risk.

The link between diabetes and heart disease is well-documented \cite{diabetesuk}. Our analysis supports this connection, as the absence of heart disease ($do(HeartDisease=0)$) decreases diabetes risk, whereas its presence ($do(HeartDisease=1)$) significantly increases it. This finding underscores the interconnected nature of these conditions, highlighting the need for integrated healthcare strategies.

\subsection{Counterfactuals}\label{subsec:counterfactual}
Counterfactuals are alternative scenarios or outcomes that could have occurred if certain events or variables had been different. In healthcare, they can be used to explore what would have happened if a different course of action was taken or if certain variables were changed. Counterfactual statements enable clinicians to investigate the probabilities associated with an 'alternate reality'. They distinguish between the indicative, which is the factual situation (events that have actually happened), and the subjunctive (events that could have occurred in an alternate reality). In this study we investigated two distinct counterfactual scenarios to assess the impact of clinician insights on our model which have been explained in Section \ref{reordering} and \ref{statements}.
\subsubsection{Reordering Variables Based on Domain Knowledge} \label{reordering}
We investigate the impact of modifying variable order within the PCF framework to illustrate its potential benefits as a proof of concept, rather than an implementation of clinician-directed decisions. While the CBN provides a robust foundational structure, our intent is to demonstrate how the model could be enhanced by integrating clinician-informed causal relationships. The core principle of our model is adaptability; it combines empirical foundations provided by the CBN with potential clinical insights. While the CBN's causal structure establishes the initial framework, we explore how clinician adjustments to the variable order might impact predictive precision.

By granting clinicians the ability to adjust variable sequences, we create an inclusive environment where domain-specific knowledge can shape and refine the model's architecture. This approach aims to balance the structured scaffolding provided by the CBN with the nuanced insights derived from clinical acumen, ultimately enriching the model's interpretability and operational efficacy in real-world healthcare settings. This fosters the exploration of counterfactual scenarios, assessing the impact of incorporating clinician insights on model performance and decision-making outcomes.

Soliciting specific feedback on variables, assumptions, and potential causal relationships enhances the interpretability, relevance, and trustworthiness of our model, ensuring alignment with clinical expertise and practice. Through this refinement, we navigate diverse scenarios or "what-if" queries related to the model's operation under distinct conditions, including modifications of causal trajectories informed by clinical expertise. 

\begin{figure*}[ht]
\centering
 \includegraphics[width=16cm]{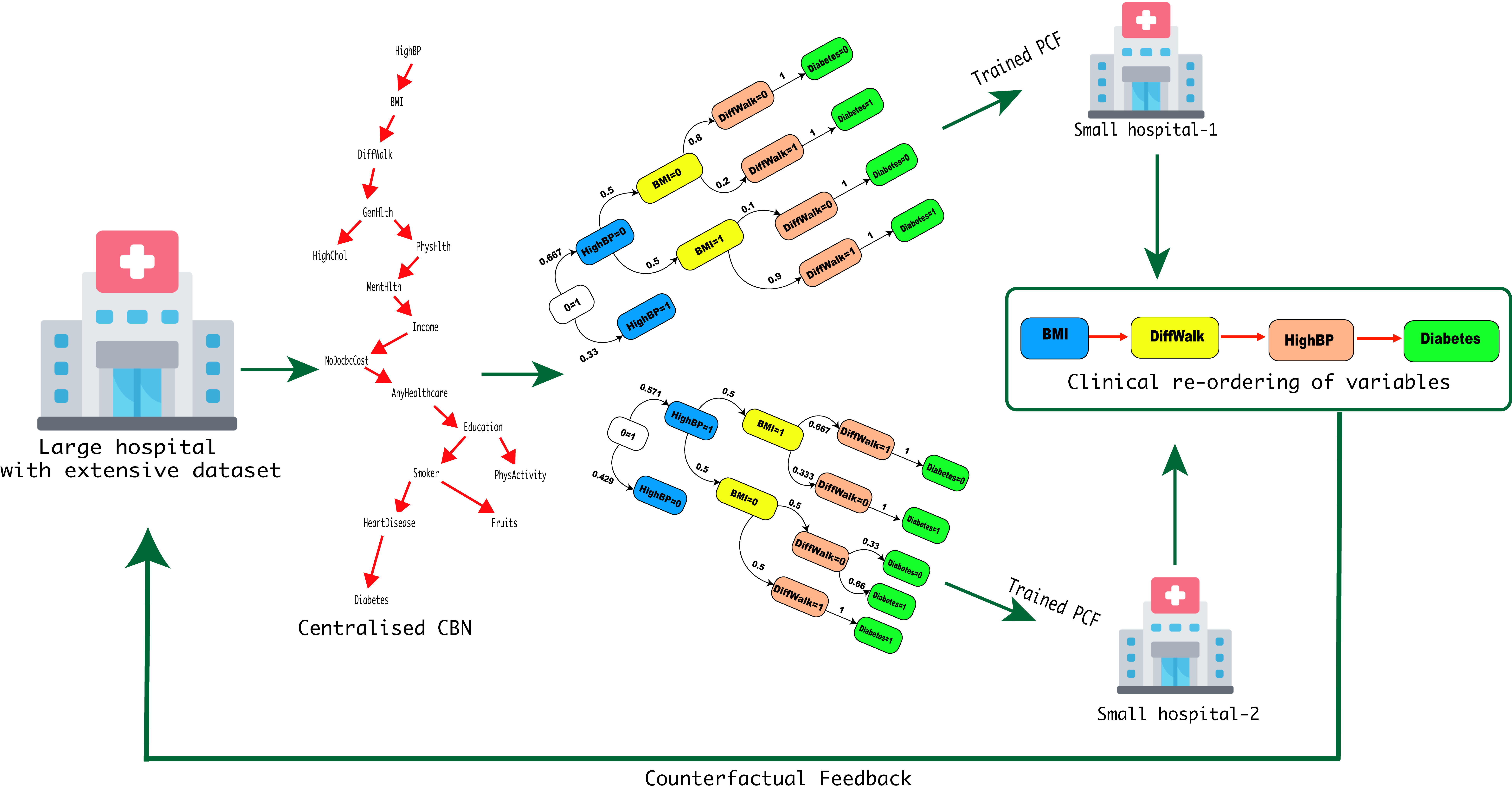}
    \caption{The process of developing and sharing pre-trained PCFs by large hospitals with extensive datasets. The feedback loop illustrates how re-ordering variables integrates clinical insights into the Centralised CBN, enhancing collaborative decision-making in healthcare. }
    \label{fig:feedbck}
    
\end{figure*}

Figure \ref{fig:feedbck} illustrates the process by which large hospitals, equipped with extensive datasets, can develop CBNs to represent causal relationships. These hospitals can then create pre-trained PCF models. These models can be securely shared with smaller hospitals, facilitating efficient knowledge transfer and collaborative decision-making in healthcare. The figure also demonstrates how re-ordering variables provides feedback to the Centralised CBN, showcasing the integration of clinical insights into the model's development.
\paragraph{\textbf{MIMIC-IV:}}
The initial variable order provided by the CBN positioned "Diagnosis\_2" earlier in the sequence, suggesting its early influence on the outcome variable. However, the counterfactual adjustment hypothesised that strategically reordering the variables might enhance the model's capacity to learn causal relationships and predict los more accurately. This adjustment reflected the potential causal flow where "Diagnosis\_2" is informed by preceding laboratory tests. Therefore, we reordered the variables to place "Diagnosis\_2" later in the sequence, just before the outcome variable (los).

Table \ref{tab:mimiccounter_results} summarises the performance of the PCF and PTree methods following the variable order change. The table reveals that while the rearrangement did not alter the accuracy of the PTree model, it resulted in a slight decrease in PCF performance. This outcome might be attributed to the sensitivity of the PCF model to variable order, as it relies on an intricate interplay of causality among variables. The reordering may have disrupted previously established dependencies, highlighting the complex interactions inherent in the data. Such changes underline the importance of carefully considering the sequence of variables in models sensitive to causal relationships.

\begin{table}
\small
  
    \caption{Results using SMOTE for MIMIC-IV after changing the order}
    \label{tab:mimiccounter_results}
    \begin{tabular}{lllll}

        \hline
        Algorithm     & Accuracy & Specificity  & Sensitivity & AUC-ROC  \\
        \hline
        \\
        
        PTree & 80.29 & 91.11 & 40.06 & 65.59 \\
        PCF & 72.43 & 78.33 & 50.50 & 64.41\\
        \hline
    \end{tabular}
\end{table}

\paragraph{\textbf{Framingham Heart Data:}}
The original variable order provided by the CBN included 'BPMeds', 'prevalentHyp', 'heartRate', 'prevalentStroke', 'diabetes', 'sysBP', 'totChol', 'glucose', 'diaBP', 'BMI', 'education', 'currentSmoker', 'cigsPerDay', and 'TenYearCHD'. Subsequently, a counterfactual adjustment was made, rearranging certain variables to create a revised order: 'BPMeds', 'prevalentHyp', 'diabetes', 'glucose', 'heartRate', 'sysBP', 'diaBP', 'BMI', 'education', 'totChol', 'prevalentStroke', 'currentSmoker', 'cigsPerDay', and 'TenYearCHD'

Table \ref{tab:framcounter_results} presents the performance evaluation results for the PCF and PTree methods following the variable order modification. The analysis indicates almost similar performance for both the PCF and PTree method (slightly lower than original). This slight decrease can be attributed to the fact that the original order might have implicitly captured relevant relationships for CHD prediction better than the counterfactual order.

\begin{table}
\small
  
    \caption{Results using SMOTE for framingham data after changing the order}
    \label{tab:framcounter_results}
    \begin{tabular}{lllll}

        \hline
        Algorithm     & Accuracy & Specificity  & Sensitivity & AUC-ROC  \\
        \hline
        \\
        
        PTree  & 64.98 & 69.40 & 40.31 & 54.85 \\
        PCF & 66.39 & 69.12 & 51.16 & 60.14\\
        \hline
    \end{tabular}
\end{table}
\paragraph{\textbf{Diabetes:}}
The original order of variables provided by the CBN was as follows: 'HighBP', 'BMI', 'DiffWalk', 'GenHlth', 'PhysHlth', 'HighChol', 'MentHlth', 'Income', 'NoDocbcCost', 'AnyHealthcare', 'Education', 'Smoker', 'PhysActivity', 'HeartDiseaseorAttack', 'Fruits', 'Diabetes\_binary'. The counterfactual order maintained the overall structure but changed the position of a few variables and the new order became: 'GenHlth','BMI','DiffWalk','PhysHlth', 'PhysActivity','HighBP', 'HighChol', 'MentHlth', 'Education', 'Income', 'NoDocbcCost', 'AnyHealthcare', 'Smoker', 'HeartDiseaseorAttack', 'Fruits', 'Diabetes\_binary'. 

Table \ref{tab:diabcounter_results} summarises the accuracy of PCF and PTree methods, after the change in variable order. As the table shows, reordering the variables led to an increase in the model accuracy for PCF as well as for PTree. However, AUC-ROC seems to have decreased in both.
\begin{table}
\small
  
    \caption{Results using ADASYN for diabetes data after changing the order}
    \label{tab:diabcounter_results}
    \begin{tabular}{lllll}

        \hline
        Algorithm     & Accuracy & Specificity  & Sensitivity & AUC-ROC  \\
        \hline
        \\
        
        PTree  & 73.29 & 78.93 & 38.14 & 58.54 \\
        PCF & 73.71 & 75.45 & 62.88 & 69.17\\
        \hline
    \end{tabular}
\end{table}

\subsubsection{Counterfactual Statements for Specific Datasets}\label{statements} 
\begin{figure*}[ht]
\centering
 \includegraphics[width=10cm]{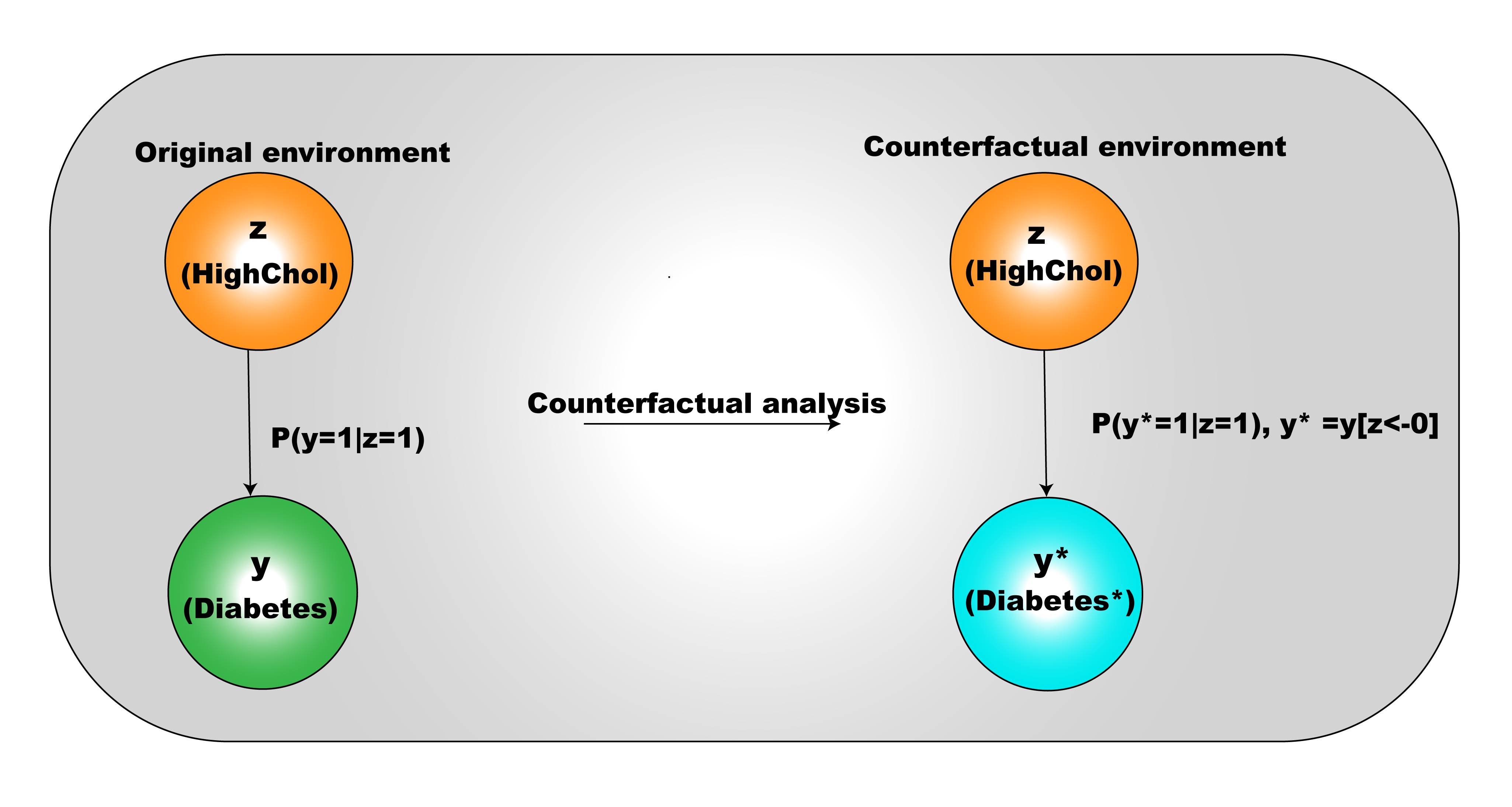}
    \caption{Visualisation of Counterfactual statements}
    \label{fig:counter}
    
\end{figure*}

Counterfactual analysis within PCF allows the exploration of different paths a stochastic process might have taken. It examines "what-if" scenarios where specific variables take on different values than in reality. This technique evaluates conditional probabilities of the form $P(A_C\mid B)$, representing representing the probability of event $A$ occurring under the counterfactual assumption that event $C$ is true, given that event $B$ is factual (observed). Here, $A_C$ denotes the subjunctive event A under the counterfactual assumption that the event C has occurred (i.e. a potential response), and B is the indicative (i.e. factual) assumption. A counterfactual PCF is constructed by modifying a reference model through factual statements, and then spawning a new variable scope from a counterfactual modification, formalised as an intervention. The intervention effectively resets the state of downstream variables by replacing their existing values with their initial unbound state. The computation of counterfactuals can be achieved algorithmically, as outlined in \cite{genewein2020algorithms}. While the power of counterfactual explanations lies in their ability to explore alternative realities, computational tractability and interpretability necessitate a focus on feature sparsity. In simpler terms, it is more computationally efficient and interpretable to manipulate a limited number of key features within the model.  This prioritises generating actionable insights by focusing on interventions that are realistic and have the potential to be implemented in practice. The burgeoning field of counterfactual explainability \cite{verma2022counterfactual} emphasises the critical importance of imposing constraints on the types of interventions considered. These constraints ensure that counterfactual explanations are not only theoretically robust but also practically applicable in real-world decision-making contexts. Guided by this principle, we carefully selected features for each dataset that offer the most valuable insights into how targeted interventions can influence predicted outcomes. This selection process was heavily informed by their established presence and significance in the existing literature. Incorporating well-researched variables enhances the relevance and reliability of our counterfactual analyses, ultimately improving the practical utility of our findings.
Figure \ref{fig:counter} presents a visual representation of counterfactual analysis.
\paragraph{\textbf{MIMIC-IV:}}
This section explores the factors influencing the LOS in the ICU using counterfactual analysis. Specifically, we examine the conditional probability of a patient requiring an extended ICU stay ($los =1$, i.e., more than 4 days). By employing counterfactual explanations, we investigate hypothetical scenarios where certain vital signs or laboratory values are altered. This allows us to assess the impact of these changes on the probability of an extended ICU stay. Features were chosen based on their potential to yield valuable insights into the determinants of prolonged ICU stays.

\begin{itemize}

\item[a)]\textbf{Heart Rate: }We analysed the effect of heart rate by comparing the baseline probability ($P(los = 1 | heart\_rate = 0)$) for patients with low heart rate (0) to the counterfactual scenario where their heart rate is normal (1). The counterfactual probability ($P(los* = 1 | heart\_rate = 0), los* = los[heart\_rate <- 1]$) suggests a decrease in the likelihood of extended ICU stay when the heart rate is normal.
\item[b)]\textbf{Saturation: }Similarly, we examined the effect of oxygen saturation by comparing baseline probability of ($P(los = 1 | saturation = 3)$) for patients with very low oxygen saturation (3) to the counterfactual scenario with normal oxygen saturation (0). The counterfactual probability ($P(los* = 1 | saturation = 3), los* = los[saturation <- 0]$) indicates that there is a marginal difference in the probability of an extended ICU stay when the patient's saturation level is normal.
\item[c)]\textbf{Glucose and Urea Nitrogen: }We further analysed the role of glucose and Urea Nitrogen levels. Interestingly, for both high glucose ($P(los = 1 | Glucose = 2)$) and high Urea Nitrogen ($P(los = 1 | Urea Nitrogen = 2)$), the counterfactual scenarios with normal levels (glucose = 1 and Urea Nitrogen = 1, respectively) showed slightly increased probabilities of extended ICU stay ($P(los* = 1 | Glucose = 2), los* = los[Glucose <- 1]$ and $P(los* = 1 | Urea Nitrogen = 2), los* = los[Urea Nitrogen <- 1]$).  Thus, the probability of extended ICU stay may be increased even if the patient had normal glucose and urea nitrogen levels.
\item[d)]\textbf{Temperature: }Finally, we explored the influence of body temperature. The baseline probability for extended ICU stay with high fever ($P(los = 1 | temperature = 2) $) was compared to the counterfactual scenario with normal temperature (0). This resulted in a minor decrease in the probability ($P(los* = 1 | temperature = 2), los* = los[temperature <- 0]$) which suggests a slight decrease in the likelihood of extended ICU stay, if the patient had normal body temperature.
\end{itemize}

\begin{figure*}[ht]
\centering
 \includegraphics[width=13cm]{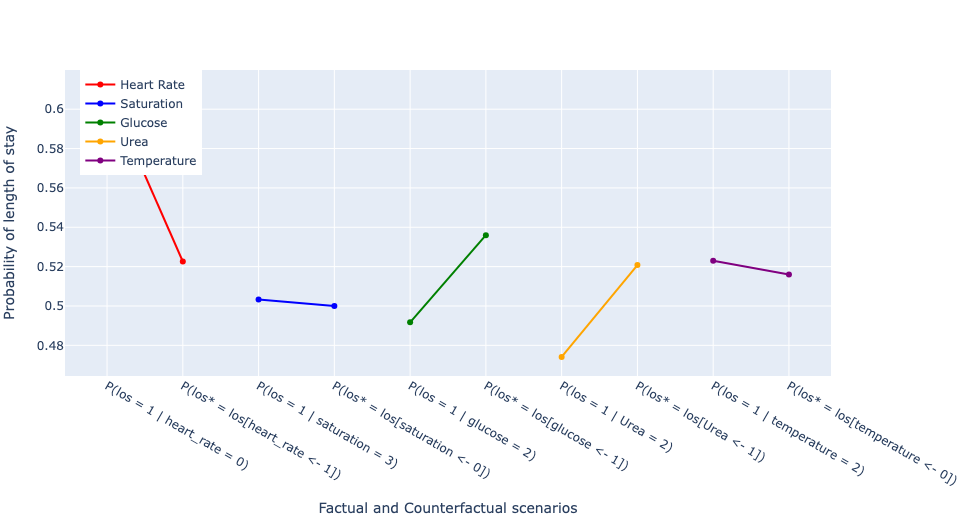}
    \caption{Line plots showing the probability of ICU stay exceeding 4 days under factual and counterfactual scenarios for various health variables.}
    \label{fig:los_counter}
    
\end{figure*}

Figure \ref{fig:los_counter} illustrates the probability of a patient remaining in the ICU for more than four days ($los = 1$) under various counterfactual scenarios involving five different variables: Heart Rate, Saturation, Glucose, Urea, and Temperature. Each line in the plot represents one of these variables, with the x-axis displaying the factual and counterfactual scenarios and the y-axis showing the probability values. This figure provides insights into how alterations in these variables influence the likelihood of an extended ICU stay.
\paragraph{\textbf{Framingham Data:}}
This data offers a wealth of information on cardiovascular risk factors. To leverage counterfactual explanations effectively, we strategically select the features with high explanatory potential for predicting CHD.
\begin{itemize}

\item[a)]\textbf{BMI: } We began our analysis by examining the impact of Body Mass Index (BMI) on the likelihood of developing CHD. Starting with a BMI of 2 (High), indicative of CHD, we delved into counterfactual scenarios to explore the probability of CHD had the patient possessed a normal BMI (0). The baseline probability $P(CHD = 1 | BMI = 2)$ represents the likelihood of CHD under the existing BMI condition. Interestingly, the counterfactual probability $P(CHD* = 1 | BMI = 2), CHD* = CHD[BMI <- 0]$ reveals that even with a normal BMI, the risk of CHD might still remain relatively high. 
\item[b)]\textbf{Systolic Blood Pressure (sysBP) and Diastolic Blood Pressure (diaBP):} 
We investigated the influence of blood pressure measurements (systolic pressure, or sysBP, and diastolic pressure, or diaBP) of patients on CHD prevalence. Individuals with high blood pressure (represented by a score of 3 for both sysBP and diaBP) were found to have a higher chance of having CHD ($P(CHD = 1 | sysBP = 3)$ and $P(CHD = 1 | diaBP = 3)$). We then considered a hypothetical scenario: what if these patients with high blood pressure had normal values instead (sysBP = 1 and diaBP = 1)? The corresponding probabilities,  ($P(CHD* = 1 | sysBP = 3), CHD* = CHD[sysBP <- 1]$ and  $P(CHD* = 1 | diaBP = 3), CHD* = CHD[diaBP <- 1]$), show how lowering blood pressure could potentially decrease the risk of CHD.
    
\item[c)]\textbf{Total Cholesterol (totChol): }We explored the potential influence of total cholesterol (totChol) on the development of CHD using counterfactual analysis. In the factual scenario, we assessed patients based on their actual totChol (totChol = 3). The baseline probability was determined as $P(CHD = 1 | totChol = 3)$. In the counterfactual scenario, we posed the question: what if these patients with high totChol had normal values (totChol = 0)?. The resulting probability $P(CHD* = 1 | totChol = 3), CHD* = CHD[totChol <- 0]$ suggests that maintaining normal total cholesterol levels might be beneficial for reducing CHD risk.
    
\item[d)]\textbf{Cigarettes per day (cigsPerDay): }Finally, we explored the influence of smoking. In the factual scenario, we examined patients based on their actual cigarette consumption (cigsPerDay = 3). The baseline probability was determined as $P(CHD = 1 | cigsPerDay = 3)$. The counterfactual scenario asks, "what if" these patients who smoke heavily cigsPerDay = 3($\geq$ 11 cigarettes/day) had never smoked (cigsPerDay = 0)?. The resulting probability $P(CHD* = 1 | cigsPerDay = 3), CHD* = CHD[cigsPerDay <- 0]$  suggests that quitting smoking could be beneficial for reducing CHD risk.
\end{itemize}

\begin{figure*}[ht]
\centering
 \includegraphics[width=13cm]{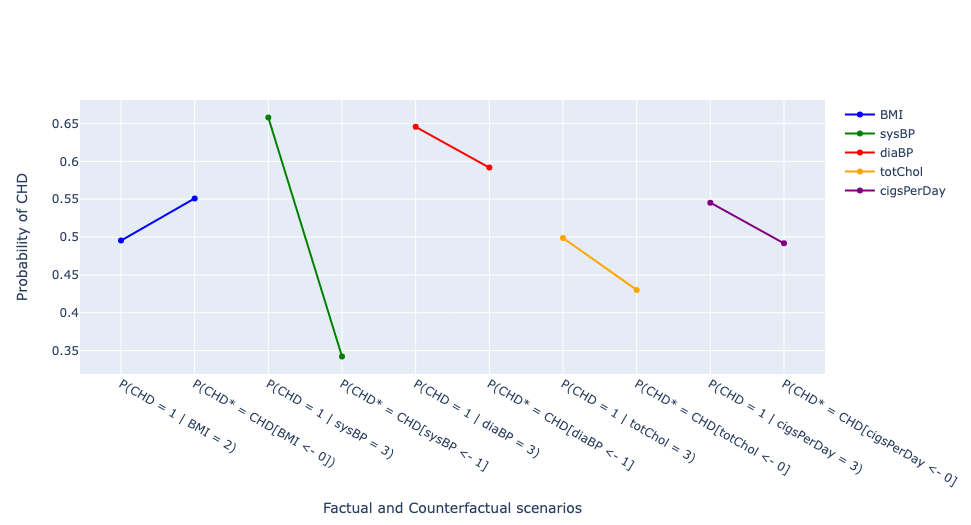}
    \caption{Probability distribution of TenYearCHD for factual and counterfactual scenarios for various variables.}
    \label{fig:fram_counter}
    
\end{figure*}

\begin{figure*}[ht]
\centering
 \includegraphics[width=13cm]{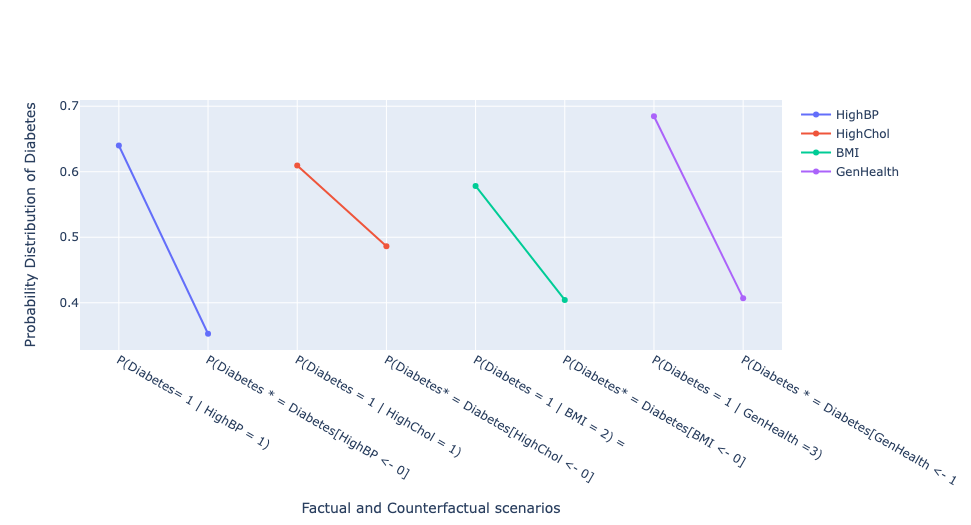}
    \caption{Probability Distribution of Diabetes for factual and counterfactual scenarios for various variables.}
    \label{fig:dia_counter}
    
\end{figure*}

Figure \ref{fig:fram_counter} illustrates the line plots generated to explore the distribution of risk factors for CHD using counterfactual analysis. It can be observed that the counterfactual scenarios pertaining to sysBP, diaBP, totChol and cigsPerDay potentially change the probability of $CHD =1$.
\paragraph{\textbf{Diabetes Data:}}  
This section explores the application of counterfactual explanations within the diabetes dataset. By strategically selecting features, we focus on identifying modifiable risk factors. 
\begin{itemize}
\item[a)]\textbf{HighBP: }We examined the relationship between HighBP and the prevalence of diabetes. In the actual scenario, patients were evaluated based on their recorded blood pressure levels (HighBP = 1). The baseline probability was calculated as $P(Diabetes\_binary = 1 | HighBP = 1)$. However, in the hypothetical scenario, we posed the question: what if these patients with high blood pressure had normal blood pressure (HighBP = 0)? The resulting probability $P(Diabetes\_binary* = 1 | HighBP = 1), Diabetes\_binary * = Diabetes\_binary [HighBP <- 0]$, suggests a potential benefit of maintaining normal blood pressure to reduce the risk of diabetes. This is in accordance to the literature which states that Blood pressure control is just as important as glycemic control \cite{mushlin2009decision}.
\item[b)]\textbf{HighChol: }We examined the influence of high cholesterol (HighChol) on the prevalence of Diabetes. In the factual scenario, patients were evaluated based on their actual cholesterol measurements (HighChol = 1). The baseline probability was computed as $P(Diabetes\_binary = 1 | HighChol = 1)$. However, in the counterfactual scenario, we considered: what if these patients with high cholesterol had normal levels (HighChol = 0)?. The resulting probability $P(Diabetes\_binary* = 1 | HighChol = 1), Diabetes\_binary * = Diabetes\_binary [HighChol <- 0]$ highlights the potential advantage of normalising cholesterol levels in mitigating the risk of diabetes.
\item[c)]\textbf{BMI: }We explored the impact of BMI on diabetes prevalence using counterfactual analysis. In the factual scenario, we assessed patients based on their actual BMI (BMI = 2). The baseline probability was $P(Diabetes\_binary = 1 | BMI = 2)$. The counterfactual scenario investigated the hypothetical scenario where patients with high BMI (BMI = 2) had a normal BMI (BMI = 0). We aimed to determine the impact of this hypothetical change on diabetes risk. The resulting probability, ($P(Diabetes\_binary* = 1 | BMI = 1), Diabetes\_binary * = Diabetes\_binary [BMI <- 0]$), suggests a potential benefit of maintaining a healthy weight (represented by normal BMI) in reducing diabetes risk.
\item[d)]\textbf{GenHealth: }This study investigated the link between a patient's overall health (GenHealth) and their risk of developing diabetes using counterfactual analysis.  The indicative premise is that the patient has poor health (GenHealth =3), and the subjunctive (counterfactual) premise is if the patient has excellent health (GenHealth =1) in an alternate reality. The baseline probability was determined as $P(Diabetes\_binary = 1 | GenHealth =3)$.  The probability resulting from the subjunctive premise, $P(Diabetes\_binary* = 1 | GenHealth =3), Diabetes\_binary * = Diabetes\_binary [GenHealth <- 1]$ illustrates the potential benefit of maintaining the overall well-being so as to reduce the risk of diabetes.
\end{itemize}
Figure \ref{fig:dia_counter} illustrates the change in $P(Diabetes = 1)$ as a result of the counterfactual statements described.

\section{Conclusion}\label{conclusion}
This study introduces the Probabilistic Causal Fusion (PCF) framework, which combines Causal Bayesian Networks (CBNs) and Probability Trees (PTrees) to enhance healthcare decision-making. This innovative approach harnesses the causal structure learned by the CBN to establish the foundational framework of the PTree. This synergy yields a three-fold benefit: (1) it captures the inherent causal relationships within the data, leading to a more robust understanding of the factors influencing outcomes, (2) it facilitates the incorporation of domain knowledge through counterfactual analysis, empowering clinicians to integrate their expertise into the model, and (3) it facilitates the creation of a centralised repository of causal knowledge across institutions. This fosters collaboration, knowledge exchange, and continuous improvement in healthcare delivery.

Rigorous validation using three real-world medical datasets demonstrates that the proposed methodology achieves prediction performance on par with established models. However, its true strength lies in its ability to surpass mere prediction and empower clinicians. Unlike traditional machine learning methods, this framework facilitates the exploration of hypothetical interventions and counterfactual scenarios through counterfactual analysis.

This enhanced functionality translates into a more comprehensive toolkit for healthcare professionals. It enables them to not only predict patient outcomes with established accuracy but also to simulate the potential effects of treatment strategies and analyse the impact of alternative scenarios. This deeper understanding of risk factor interactions, intervention effects, and the flexibility to explore variable ordering significantly improves clinical decision-making, ultimately leading to optimised patient care.

A key strength of this approach is its dual applicability at the individual and population levels. Clinicians can leverage this framework to gain insights into broader population trends while simultaneously exploring personalised treatment options for specific patients through counterfactual analysis. This versatility empowers healthcare professionals to tailor their decision-making to the unique circumstances of each patient while simultaneously informing clinical best practices for the entire population.

Our study suggests that this approach holds significant promise for evidence-based clinical decision-making. However, further exploration is needed to address certain limitations. Optimising computational efficiency, especially for large datasets, is crucial for broader applicability. While this study focused on specific medical domains, future research should investigate the framework's generalisability to a wider range of healthcare settings. Incorporating clinical expertise in selecting variables for counterfactual and interventional analysis is essential. Clinicians' insights can refine the methodology and enhance its practical utility by ensuring the system uses the most relevant and useful data.

Addressing these limitations and broadening the scope of applications, including genomic data analysis, will further demonstrate the framework's potential to advance healthcare. Building on this approach can lead to the development of more effective and transparent tools that enhance patient care. Such tools have the potential to support evidence-based clinical decision-making and contribute to a more efficient and impactful healthcare system overall.

 \section*{CRediT Authorship Contribution Statement}
\textbf{Sheresh Zahoor:} Conceptualization, Methodology, Data
 curation, Visualization,  Formal analysis, Investigation,
 Resources, Software, Validation, Writing – original draft, Writing – review and editing.
\textbf{Pietro Liò:} Conceptualization, Investigation,
 Resources, Supervision, Validation, Writing – review and editing.
\textbf{Gaël Dias:} Conceptualization, Supervision, Validation, Writing – review and editing.
\textbf{Mohammed Hasanuzzaman:} Conceptualization, Supervision, Validation, Writing – review and editing.
 \section*{Declaration of Competing Interest}
 The authors declare that they have no known competing financial
interests or personal relationships that could have appeared to influence
 the work reported in this paper.
\section*{Acknowledgments}
This work has been jointly supported by Research Ireland [Grant number 18/CRT/6222] and the European Union’s Horizon 2020 research and innovation programme[Grant agreement number 101017385]. Additionally, for the purpose of Open Access, the author has applied a CC BY public copyright licence to any Author Accepted Manuscript version arising from this submission.

\bibliographystyle{elsarticle-num} 
\bibliography{primary}
\end{document}


\begin{enumerate}
\item\textbf{los (length of stay):} We exclude patients with an LOS greater than 21 days to avoid extremely long stays, and less than one day since we consider data collected in the first 24 h for modeling. LOS was categorised into short and long stay which was further encoded numerically as 0 and 1.
\item\textbf{heart\_rate :} Heart rate was categorised into Low (<60 beats per minute), Normal (60-100 beats per minute) and High (>100 beats per minute). These were further encoded as 0,1 and 2.
\item\textbf{respiration :} Respiration rates falling within the range of 12 to 20 breaths per minute were categorized as 'Normal', those below 12 breaths per minute were labeled as 'Low', and rates exceeding 20 breaths per minute were classified as 'Abnormal'. This was encoded as 0, 1 and 2.
\item\textbf{saturation :}To categorize oxygen saturation levels into clinically meaningful groups, we employed a categorization function. Oxygen saturation readings were classified as follows: 'Normal reading' for readings at or above 95\%. 'Insufficient oxygen levels' for readings between 91\% and 94\%. 'Low blood oxygen level' for readings between 85\% and 90\%. 'Very low oxygen levels (Hypoxemia)' for readings between 80\% and 84\%. 'Severe Hypoxemia' for readings between 67\% and 79\%. 'Cyanosis' for readings below 67\%. This was further encoded numerically as 0,1,2,3,4 and 5 respectively.

\item\textbf{temperature :}Body temperatures falling within the range of 95°F to 100.4°F (equivalent to 35°C to 38°C) were classified as 'Normal'. Temperatures ranging from 100.4°F to 102.2°F (equivalent to 38°C to 39°C) were categorized as 'Mild Fever', while temperatures equal to or exceeding 102.2°F (equivalent to 39°C) were labeled as 'High Fever'. Additionally, temperatures below 95°F (equivalent to 35°C) were identified as 'Hypothermia'. This was further encoded as 0, 1, 2 and 3

\item\textbf{Hematocrit :}In our analysis, we implemented a gender-specific classification function to categorize hematocrit levels into clinically relevant groups. For individuals identified as male, hematocrit levels below 41\% were categorized as 'Low', while levels equal to or exceeding 55\% were classified as 'High'. Hematocrit levels falling within the range of 40\% to 54\% were deemed 'Normal' for this group. Conversely, for individuals identified as female, hematocrit levels below 35\% were classified as 'Low', levels equal to or exceeding 49\% were categorized as 'High', and levels falling within the range of 36\% to 48\% were deemed 'Normal'. 
\item\textbf{Hemoglobin :} We employed a gender-specific approach to categorize hemoglobin levels into clinically significant groups. For individuals identified as male , hemoglobin levels were categorized as follows: 'Low' if below or equal to 13 g/dL, 'High' if equal to or exceeding 19 g/dL, and 'Normal' if falling within the range of 14 g/dL to 18 g/dL. Conversely, for individuals identified as female, hemoglobin levels were classified as 'Low' if below or equal to 11 g/dL, 'High' if equal to or exceeding 17 g/dL, and 'Normal' if falling within the range of 12 g/dL to 16 g/dL. This was further encoded numerically as 0, 1 and 2
\item\textbf{Platelet Count :} 
Platelet counts below $150 \times 10^3/μL$ were classified as 'Low'. Counts falling within the range of $150 \times 10^3/μL$ to $400 \times 10\^3/μL$ were categorized as 'Normal', indicating a typical physiological range. Platelet counts exceeding $400 \times 10^3/μL$ were designated as 'High'. This was further encoded numerically as 0, 1 and 2.
\item\textbf{White Blood Cells :} We established distinct categories to delineate different levels of WBC counts, reflecting potential health implications. WBC counts below a threshold of $4 \times 10^3/μL $ were categorized as 'Low', potentially indicative of leukopenia or immunocompromised states. Counts falling within the range of $4 \times 10^3/μL$ to $10 \times 10^3/μL$ were classified as 'Normal', representing the typical physiological range for WBC counts in adults. WBC counts surpassing $10 \times 10^3/μL$ were designated as 'High'. This was further encoded numerically as 0, 1 and 2.
\item\textbf{Anion Gap :} Anion gap values below 8 mmol/L were categorized as 'Low', potentially indicating conditions such as hypoalbuminemia or certain metabolic disorders. Values falling within the range of 8 mmol/L to 20 mmol/L were classified as 'Normal', representing the typical physiological range for anion gap in adults. Anion gap values exceeding 20 mmol/L were designated as 'High'. This was further encoded numerically as 0, 1 and 2.
\item\textbf{Bicarbonate :} Bicarbonate levels below 22 mmol/L were categorized as 'Low', which may indicate metabolic acidosis or other acid-base disturbances. Levels falling within the range of 22 mmol/L to 32 mmol/L were classified as 'Normal', representing the typical physiological range for bicarbonate in adults. Bicarbonate levels exceeding 32 mmol/L were designated as 'High'.This was further encoded numerically as 0, 1 and 2.
\item\textbf{Chloride :}Chloride levels below 96 mmol/L were categorized as 'Low', which may indicate conditions such as metabolic alkalosis or certain electrolyte imbalances. Levels falling within the range of 96 mmol/L to 108 mmol/L were classified as 'Normal', representing the typical physiological range for chloride in adults. Chloride levels exceeding 108 mmol/L were designated as 'High'. This was further encoded numerically as 0, 1 and 2.
\item\textbf{Creatinine :} Creatinine levels below 0.5 mg/dL were categorized as 'Low', which may indicate conditions such as muscle wasting or reduced kidney function. Levels falling within the range of 0.5 mg/dL to 1.2 mg/dL were classified as 'Normal', representing the typical physiological range for creatinine in adults. Creatinine levels exceeding 1.2 mg/dL were designated as 'High'. This was further encoded numerically as 0, 1 and 2.
\item\textbf{Glucose: }Glucose levels below 70 mg/dL were categorized as 'Low', which may indicate hypoglycemia and potentially pose risks such as impaired cognitive function or loss of consciousness. Levels falling within the range of 70 mg/dL to 100 mg/dL were classified as 'Normal', representing the typical physiological range for fasting blood glucose in adults. Glucose levels exceeding 100 mg/dL were designated as 'High'. This was further encoded numerically as 0, 1 and 2.
\item\textbf{Sodium :} Sodium levels below 135 mmol/L were categorized as 'Low', which may indicate hyponatremia and potentially pose risks such as altered mental status or neurological symptoms. Levels falling within the range of 135 mmol/L to 147 mmol/L were classified as 'Normal', representing the typical physiological range for serum sodium in adults. Sodium levels exceeding 147 mmol/L were designated as 'High'.This was further encoded numerically as 0, 1 and 2.
\item\textbf{Potassium :} Potassium levels below 3.3 mmol/L were categorized as 'Low', which may indicate hypokalemia and potentially pose risks such as muscle weakness or cardiac arrhythmias. Levels falling within the range of 3.3 mmol/L to 5.1 mmol/L were classified as 'Normal', representing the typical physiological range for serum potassium in adults. Potassium levels exceeding 5.1 mmol/L were designated as 'High'.This was further encoded numerically as 0, 1 and 2.
\item\textbf{Red Blood Cells :}RBC counts below 4.6 million cells per microliter (μL) were categorized as 'Low', which may indicate conditions such as anemia or blood loss. Counts falling within the range of 4.6 million cells/μL to 6.2 million cells/μL were classified as 'Normal', representing the typical physiological range for RBC counts in adults. RBC counts exceeding 6.2 million cells/μL were designated as 'High'.This was further encoded numerically as 0, 1 and 2.
\item\textbf{Red Cell Distribution Width (RDW) :} RDW values below 10.5\% were categorized as 'Low', which may indicate certain types of anemias with uniform cell size. Values falling within the range of 10.5\% to 15.5\% were classified as 'Normal', representing the typical physiological range for RDW in adults. RDW values exceeding 15.5\% were designated as 'High'.This was further encoded numerically as 0, 1 and 2.
\item\textbf{Mean Corpuscular Hemoglobin (MCH) :} MCH values below 27 picograms (pg) were categorized as 'Low', which may indicate hypochromic anemia or iron deficiency. Values falling within the range of 27 pg to 32 pg were classified as 'Normal', representing the typical physiological range for MCH in adults. MCH values exceeding 32 pg were designated as 'High'. This was further encoded numerically as 0, 1 and 2.
\item\textbf{Mean Corpuscular Hemoglobin Concentration (MCHC) :} MCHC values below 32 grams per deciliter (g/dL) were categorized as 'Low', which may indicate hypochromic anemia or conditions associated with decreased hemoglobin concentration in red blood cells. Values falling within the range of 32 g/dL to 37 g/dL were classified as 'Normal', representing the typical physiological range for MCHC in adults. MCHC values exceeding 37 g/dL were designated as 'High'. This was further encoded numerically as 0, 1 and 2.
\item\textbf{Mean Corpuscular Volume (MCV) :}MCV values below 82 femtoliters (fL) were categorized as 'Low', which may indicate microcytic anemia or conditions associated with smaller-than-normal red blood cells. Values falling within the range of 82 fL to 98 fL were classified as 'Normal', representing the typical physiological range for MCV in adults. MCV values exceeding 98 fL were designated as 'High'. This was further encoded numerically as 0, 1 and 2.
\item\textbf{Urea Nitrogen :}Urea Nitrogen values below 6 milligrams per deciliter (mg/dL) were categorized as 'Low', which may indicate conditions such as liver disease or malnutrition. Values falling within the range of 6 mg/dL to 20 mg/dL were classified as 'Normal', representing the typical physiological range for Urea Nitrogen in adults. Urea Nitrogen values exceeding 20 mg/dL were designated as 'High'.This was further encoded numerically as 0, 1 and 2.
\item\textbf{Magnesium :}Magnesium levels below 1.6 milligrams per deciliter (mg/dL) were categorized as 'Low', which may indicate hypomagnesemia and potential risks such as muscle weakness, cardiac arrhythmias, or neuromuscular abnormalities. Values falling within the range of 1.6 mg/dL to 2.6 mg/dL were classified as 'Normal', representing the typical physiological range for magnesium in adults. Magnesium levels exceeding 2.6 mg/dL were designated as 'High'.This was further encoded numerically as 0, 1 and 2.
\item\textbf{anchor\_age :} Among ICU patients, we only select adults older than 18 years. Those aged between 18 and 29 are classified as group 0, while individuals aged 30 to 44 are categorized as group 1. Group 2 consists of individuals aged 45 to 54, while group 3 includes those aged 55 to 64. Individuals aged 65 to 74 are assigned to group 4, and those aged 75 and above are placed in group 5. 
\end{enumerate}

\section{Methodology}

This study presents a novel pipeline that merges the methodological advantages of CBNs and PTrees for predictions, interventions and counterfactuals. The traditional iterative process of constructing PTrees involves two fundamental aspects: the judicious selection and arrangement of features guided by domain knowledge (deduction), and the determination of transition probabilities informed by empirical data (induction). Thus, the construction of the PTrees is done based on domain knowledge obtained from experts and relevant literature. However, this is up to the human researcher to conclude, and a different design will result in a difference in performance. The performance of the model is dependent on the design of the tree, and it is done according to the beliefs of the researcher/domain experts. While domain knowledge plays a crucial role, this process can be subjective and time-consuming. Our proposed pipeline introduces a novel approach by integrating CBNs to autonomously establish the variable order, thereby facilitating PTree construction. This integration not only enables the seamless adoption of the established PTree framework delineated in prior research \cite{genewein2020algorithms}, but also offers an automated pathway for its application.  

Additionally, a pivotal next step involves the formation of an ensemble comprising multiple PTrees, aimed at augmenting both predictive accuracy and performance, with a particular focus on addressing the challenges posed by complex problem scenarios.  This ensemble strategy can reduce the risk of overfitting and underfitting by balancing the trade-off between bias and variance, and the utilisation of diverse subsets of the data.
We further explain the two steps of the pipeline in \ref{CBN analysis} and \ref{PTree Construction}

\subsection{Causal Bayesian Network Analysis}
\label{CBN analysis}
Causal Bayesian Networks (CBNs) serve as a robust framework for elucidating the causal interrelations among variables within intricate datasets. In this phase of the investigation, CBNs were utilised to unveil the causal architecture of the chosen variables and ascertain the optimal sequence in which they exert influence on the outcome variable. This sequencing was deemed pivotal in the subsequent development of a proficient PTree. To accomplish this, we employed the ``target variable'' knowledge approach, as provided within the Bayesys \cite{anthony} open-source Bayesian Network Structure Learning (BNSL) system. This methodological approach prioritises structure learning to elucidate a greater number of parent nodes, conceived as potential causal factors, pertaining to the variables of interest. Although the primary aim of this approach may not be the enhancement of graph accuracy, it remains pertinent when dealing with high-dimensional real-world data characterised by limited sample size. This is attributable to the tendency of structure learning algorithms in such scenarios to yield sparse networks that inadequately capture the potential causal factors or parent nodes relevant to the targeted variables of interest.

Five distinct algorithms, namely Hill Climbing (HC) \cite{heckerman1995learning,hc}, TABU \cite{bouckaert1995bayesian}, SaiyanH \cite{saiyanh}, Model-Averaging Hill-Climbing (MAHC) \cite{constantinou2022effective}, and Greedy Equivalence Search (GES) \cite{ges} were utilised in the construction of CBNs. The selection of these algorithms was based on their capability to incorporate the target variable during model construction, accommodating diverse knowledge approaches such as direct relationships and forbidden edges \cite{constantinou2023impact}. This approach provides the advantage of integrating algorithmic insights with prior domain knowledge regarding cause-effect relationships. Consequently, this methodology facilitates a cohesive integration of both knowledge-based and algorithmic-driven strategies, thereby enhancing the robustness of the modeling process.
Subsequently, we adopted a model-averaging approach facilitated by Bayesys \cite{anthony}, which accepts multiple graphical structures of varying types as input and operates as three successive steps:
\begin{itemize}
\item[S1.] Add directed edges to the average graph, starting with the edges that occur most frequently across input graphs.
\begin{itemize}
    \item[a)] If an edge has already been added in the reverse direction, skip it.
    \item[b)] If adding an edge would create a cycle, reverse the edge and add it to the edge-set $C$ i.e., this edge-set contains the arcs that cause a cycle if added to the model-averaging graph at this step, and - instead of deleting them - it reverses them to be considered for inclusion in the model-averaging graph at Step 3.
    \end{itemize}
\end{itemize}
\begin{itemize} 
\item[S2.] Add undirected edges to the average graph, starting with the edges that occur most frequently.
\begin{itemize}
    \item[a)] Skip an edge if it has already been added as a directed edge.
     \end{itemize}
\end{itemize}
\begin{itemize} 
\item[S3.] Add directed edges from the edge-set $C$ to the average graph, starting with the edges that occur most frequently.
\begin{itemize}
    \item[a)] Skip an edge if it has already been added as an undirected edge.
    \end{itemize}
\end{itemize}

\begin{figure*}[ht]
\centering
 \includegraphics[width=14cm]{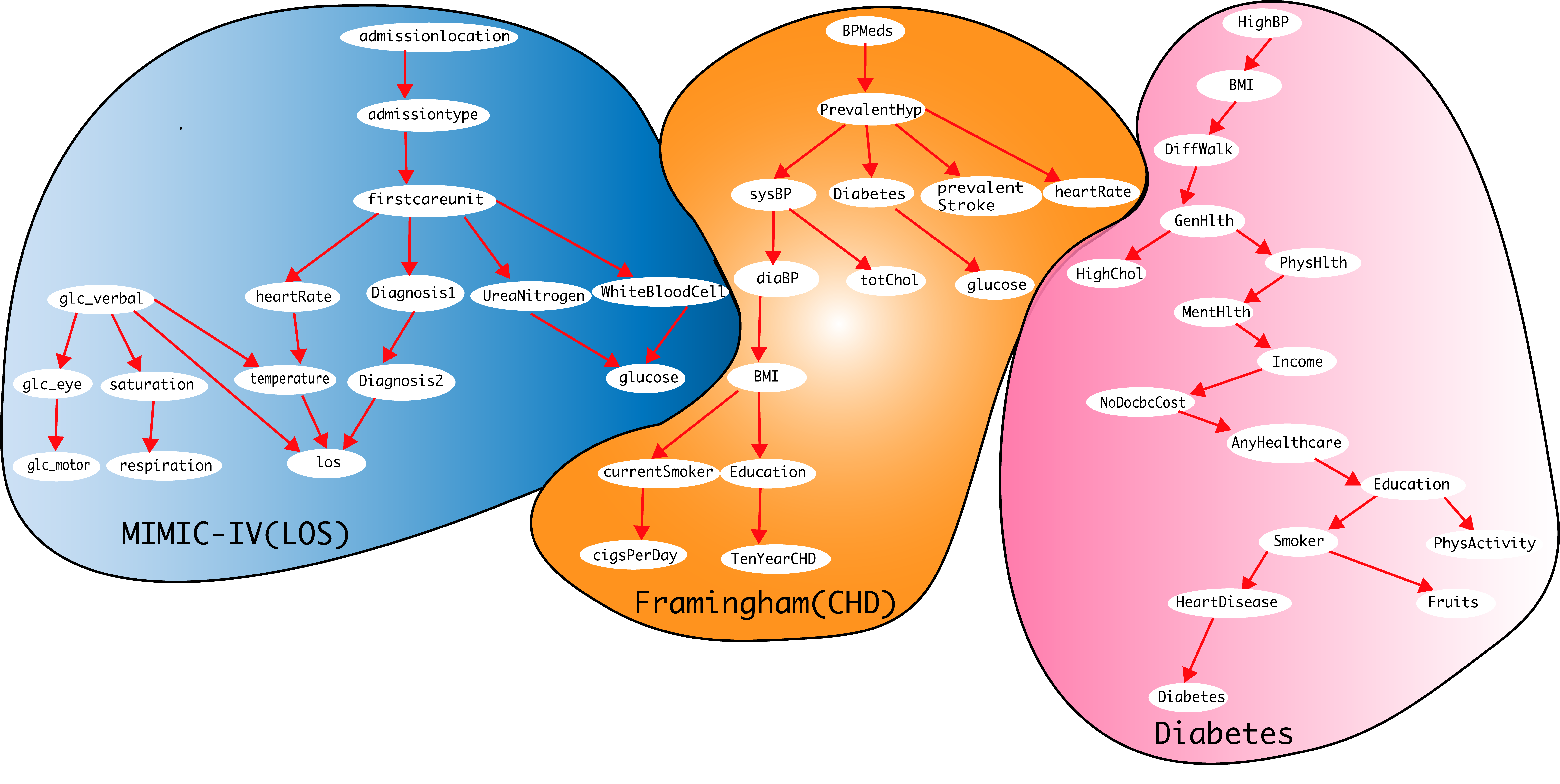}
    \caption{The model-averaging graph obtained from 5 structure learning algorithms. The red arcs depicts the variable order which is being followed for the construction of PTree. }
    \label{fig:model_average}
    
\end{figure*}

Following the construction of a model-averaging graph, the subsequent step entailed the adoption of topological order sorting to capture the sequential arrangement of variables. This sorted order was then utilised in the construction of PTrees, facilitating their development and enhancing their predictive capabilities. Figure \ref{fig:model_average} illustrates the model-average graphs generated for the three case studies, serving as a crucial foundation for constructing the PTree. The red arcs in the figure highlight the specific variable order selected for the PTree construction.

\subsection{Probability Tree Construction} 
\label{PTree Construction}
This work builds upon the existing framework of Probability Trees (PTrees) as developed in \cite{genewein2020algorithms}. However, we introduce key modifications to integrate them seamlessly within the proposed pipeline. The process commences with Algorithm \ref{alg:create_tree} (\textsc{create\_tree\_from\_data}), which outlines the construction of a PTree from a given dataset and the variable order derived from the previously learned CBN. One of our key contributions lies in the methodological refinement pertaining to the computation of transition probabilities within the PTree framework. At the root level, the algorithm underscores simplicity by partitioning data based on the prevalent value of the target attribute. This initial partitioning is ascertained through the calculation of empirical marginal probabilities for each value, as in equation \ref{eq:p_v}:
\begin{equation}
p(v) = \frac{\text{Count}(v)}{\text{Total Samples}}
\label{eq:p_v}
\end{equation}

\noindent where:
\begin{itemize}
    \item[] \(p(v)\) denotes the transition probability for a given value \(v\) of the target attribute.
    \item[] \(\text{Count}(v)\) represents the number of occurrences of value \(v\) in the target attribute.
    \item[] \(\text{Total Samples}\) represents the total number of samples in the dataset.
\end{itemize}
Moving down the tree, the model leverages the causal structure learned from the CBN to make more informed split decisions. Here, transition probabilities are calculated as the conditional probability of the current value occurring given the value of the parent node. This context-aware approach allows the tree to make more informed split decisions, potentially enhancing its predictive capabilities.
Formally, the conditional probability is expressed as:
\begin{equation}
P(X_{\text{current}} | X_{\text{parent}}) = \frac{\text{Count}(X_{\text{parent}})}{\text{Count}(X_{\text{current}}, X_{\text{parent}})}
\end{equation}

where:
\begin{itemize}
    \item[] \(P(X_{\text{current}} | X_{\text{parent}})\) represents the transition probability of the current attribute given the parent attribute.
    \item[] \(\text{Count}(X_{\text{current}}, X_{\text{parent}})\) denotes the number of occurrences of the current value given the parent value.
    \item[] \(\text{Count}(X_{\text{parent}})\) represents the total number of occurrences of the parent value.
\end{itemize}

To prevent overfitting, a pruning step is employed before incorporating a new child node. This involves evaluating the calculated conditional probability against a predefined threshold. Only nodes exceeding this threshold are added to the final tree structure, ensuring the model prioritises informative splits and ultimately strengthens its generalisation performance on unseen data. A key innovation of our proposed method is the use of ensemble learning to build a robust prediction model. This approach offers significant benefits compared to using a single PTree. Ensemble learning combines the strengths of multiple, diverse models, leading to better performance and reducing the potential weaknesses associated with single-model approaches. This collective approach enhances the reliability and accuracy of the predictions.

The ensemble learning process within our framework is implemented through the $ensemble\_probability\_trees$ function. This function employs a three-step strategy:
\begin{itemize}
    \item[1]\textbf{Strategic Data Splitting:} The data is strategically divided into batches. This step ensures that each PTree is trained on a distinct subset of the data distribution. This diversity in training data promotes the capture of a broader range of relationships within the overall dataset.
     \item[2]\textbf{Construction of Diverse PTrees:} A PTree is constructed from each data batch using Algorithm \ref{alg:create_tree} ($create\_tree\_from\_data$). This results in a collection of diverse PTrees, each potentially capturing slightly different yet complementary aspects of the data's underlying structure.
     \item[3]\textbf{Root Node Storage:} The root nodes of these individual PTrees are stored in a list ($ptrees$). These root nodes act as entry points for the prediction process within each respective tree.
\end{itemize}

By incorporating predictions from this diverse ensemble of PTrees, the model achieves a more comprehensive understanding of the complex relationships within the data. This, in turn, strengthens the foundation for the prediction task performed by the $make\_predictions$ function. This function leverages the $ptrees$ list as input and performs predictions for each PTree within the ensemble. These individual predictions are generated using modified prediction methods, as detailed in
\cite{ambags2023assisting}. An average of these individual predictions is then calculated and returned as $averaged\_predictions$. These averaged predictions represent the probability associated with a data point belonging to the negative or positive class. To categorise data points, a threshold is employed. Data points with averaged predictions exceeding this threshold are classified as belonging to the positive class. This threshold can be modified depending upon the cases we are dealing with. As an integral component of the predictive framework, Algorithm \ref{alg:conditional_probability} calculates the conditional probability of a specified class based on a defined set of input conditions. Utilising the pre-established PTree, the algorithm begins by checking for the presence of input conditions. It then systematically processes each condition using the MinCut methodology as detailed in \cite{genewein2020algorithms} to extract the relevant probabilities from the tree. These probabilities are then combined using a logical AND operation to ensure all specified conditions are met.

o address these challenges, we propose the Probabilistic Causal Fusion (PCF) framework, which integrates Causal Bayesian Networks (CBNs) with Probability Trees (PTrees). CBNs and PTrees are powerful tools for understanding complex causal patterns in healthcare data \cite{sanchez2022causal, prosperi2020causal, rajput2022causal, ambags2023assisting}. While CBNs excel at modeling causal relationships through a directed acyclic graph (DAG) structure that highlights dependencies and causal paths, PTrees offer an intuitive representation of probabilistic relationships using nodes and branches. This hierarchical structure simplifies comprehension of complex probabilistic dependencies and facilitates reasoning about uncertainty, a common aspect of healthcare decision-making \cite{genewein2020algorithms}. This complementary nature motivates the development of PCF.

While Randomised Controlled Trials (RCTs) are the gold standard for establishing causal relationships, they can be both costly and ethically challenging. The PCF framework offers a compelling alternative by matching the predictive power of benchmark models while enhancing interpretability. By integrating Causal Bayesian Networks (CBNs) with Probability Trees (PTrees), PCF uncovers complex cause-effect relationships, supports targeted interventions, and facilitates counterfactual analysis. This approach ensures high predictive accuracy and produces actionable insights, empowering clinicians with a deeper understanding of patient data and treatment options.

Moreover, the advanced decision tree methodology in PCF enables the seamless integration of diverse data sources, including clinical bedside information. CBN acts as a robust component for investigating dependencies among morbidities and assessing temporal links between comorbidities. Meanwhile, the PTree structure quantifies the strength of morbidity-comorbidity relationships across different patient cohorts, offering clinicians a comprehensive tool for informed decision-making.

The PCF framework transcends the limitations of individual models by harnessing the complementary strengths of CBNs and PTrees. By leveraging the rich causal knowledge gleaned from the CBN, PCF informs the structure of an ensemble of PTrees. This critical step ensures that the modeled probabilistic dependencies faithfully reflect the underlying causal relationships inherent in the data, a key advantage absent in traditional methods. Traditionally, constructing a PTree is an iterative and often subjective process, heavily reliant on domain experts for variable ordering \cite{ambags2023assisting}. This dependence introduces limitations:
\begin{enumerate}
\item Subjectivity: domain knowledge can be biased or incomplete, potentially leading to suboptimal PTree structures;
\item Inefficiency: the iterative process can be time-consuming, especially for complex datasets;
\item Data Inconsistency: PTrees constructed solely on expert knowledge might not fully capture the relationships present in the data.
\end{enumerate}
Furthermore, by employing an ensemble learning approach, PCF mitigates the limitations associated with single PTrees, particularly their susceptibility to overfitting.

The PCF framework not only functions as a sophisticated modeling tool but also establishes a centralised causal knowledge repository, facilitating collaboration and knowledge exchange among healthcare institutions. It offers pre-trained PCF models and their underlying causal structures, enabling smaller institutions to leverage collective expertise and develop generalisable models applicable across diverse healthcare settings. This collaborative ecosystem fosters continuous improvement as institutions refine models based on shared experiences and local data. Figure \ref{fig:illustration} summarises different steps of the proposed PCF framework.

\subsection{Interpretable ML}
IML aims to create models inherently understandable to humans, emphasising transparency from the outset \cite{bacsaugaouglu2022review}. Various IML approaches have been explored in healthcare. Kennedy et al. \cite{kennedy2021development} used EHR to train ML models predicting the 60-day risk of major adverse cardiac events in adults with chest pain, with SuperLearner Stacked Ensembling (SLSE) achieving the best performance, interpreted through Accumulated Local Effect (ALE) visualisation and variable importance ranking. Moreno-Sanchez et al. \cite{moreno2020development} developed an interpretable model for predicting heart failure survival using ensemble methods. Miran et al. \cite{miran2021model} introduced the effect score, a model-agnostic technique, to evaluate the impact of age and comorbidities on heart failure prediction. The R.ROSETTA platform \cite{garbulowski2021r} utilised rule-based methods to develop nonlinear IML models, generating interpretable predictions.
\subsection{Explainable AI}
XAI encompasses processes and methods to describe deep learning models by characterising model accuracy, transparency, and outcomes \cite{arrieta2020explainable}. Amoroso et al. \cite{amoroso2021roadmap} applied clustering and dimension reduction to identify key clinical features and design oncological therapies within an XAI framework. Uddin et al. \cite{uddin2022deep} utilised LSTM-based RNNs to locate texts depicting self-perceived depressive symptoms, employing the XAI technique LIME for model interpretability. Amorim et al. \cite{amorim2023evaluating} proposed a method for assessing saliency maps' accuracy using natural image perturbations and examining their effects on saliency model assessment metrics.
Despite advances in IML and XAI, many techniques still struggle to provide comprehensive causal insights, which are crucial for understanding data-generating processes. Clinicians require the ability to identify causes and effects, predict treatment outcomes, and explore counterfactual scenarios. Although some XAI methods pass basic causal certification, they often fail to achieve complete causal certification, limiting their generalisability and applicability in diverse clinical contexts \cite{baron2023explainable}.
\subsection{Causal Bayesian Networks and Probability Trees}
CBNs are powerful tools for modeling complex relationships in healthcare data. Rajendran et al. \cite{rajendran2020predicting} integrated CBNs with risk factor analysis in breast cancer research to enhance early detection and identify high-risk individuals. Shahmirzalou et al. \cite{shahmirzalou2023survival} used CBNs to predict survival outcomes in recurrent breast cancer, aiding personalised treatment plans. Jang et al. \cite{jang2024estimating} applied CBNs to model risks associated with radiation therapy, supporting personalised treatment decisions. Ordovas et al. \cite{ordovas2023bayesian} employed CBNs to explore relationships between cardiovascular risk factors and related conditions. Ambags et al. \cite{ambags2023assisting} proposed probabilistic fuzzy decision trees and applied them in a proof-of-concept to two real medical scenarios. Despite their simplicity and expressiveness, PTrees have received less attention compared to CBNs and structural causal models in the statistical and machine learning literature. PTrees, grounded in the principles of probability theory, represent decision-making processes through a tree structure where each node signifies a probabilistic event and branches represent conditional probabilities of subsequent events. Unlike Bayesian Networks, which use directed acyclic graphs to depict dependencies among variables, PTrees explicitly outline the sequential nature of decisions and outcomes, offering a clear and intuitive framework for modeling complex stochastic processes \cite{genewein2020algorithms}.

Healthcare decision-making requires not only accurate predictions but also a deeper understanding of the relationships between factors and their quantified impacts on patient outcomes. Traditional machine learning (ML) models excel at predicting outcomes, such as identifying patients at high risk for a condition. However, they often fall short in answering "what if" questions, such as quantifying how interventions or treatments influence outcomes. This study introduces the Probabilistic Causal Fusion (PCF) framework, which combines Causal Bayesian Networks (CBNs) with ensembles of Probability Trees (PTrees) to go beyond prediction and explore causal relationships. PCF leverages the causal order learned by the CBN to structure PTrees, enabling the quantification of factor impacts and facilitating the exploration of hypothetical interventions.

Validated on three real-world healthcare case studies—MIMIC-IV, Framingham Heart Study, and Diabetes datasets—PCF demonstrates predictive performance comparable to traditional ML benchmarks while offering additional capabilities for causal reasoning. These case studies were selected for their clinically diverse variables, allowing PCF to uncover relationships between modifiable factors.

To enhance interpretability, PCF integrates sensitivity analysis and SHapley Additive exPlanations (SHAP). Sensitivity analysis quantifies the influence of causal parameters on key outcomes, such as Length of Stay (LOS), Coronary Heart Disease (CHD), and Diabetes, while SHAP values provide transparency in the predictive modeling component by identifying the importance of individual features. These tools help bridge the gap between clinical intuition and data-driven insights, enabling a clearer understanding of the interplay between risk factors and outcomes.

By integrating causal reasoning with predictive modeling, PCF advances beyond traditional ML approaches by quantifying relationships and providing tools to simulate hypothetical scenarios. This capability allows clinicians to better understand the causal pathways underlying outcomes and supports more informed, evidence-based decision-making across diverse healthcare settings.